%% file: main.tex
\documentclass[10pt,twocolumn,letterpaper]{article}

\usepackage[pagenumbers]{cvpr} 

\input{preamble}

\usepackage[pagebackref,breaklinks,colorlinks,allcolors=cvprblue]{hyperref}

\def\paperID{VAND-7} 
\def\confName{CVPR}
\def\confYear{2025}

\title{SmartHome-Bench: A Comprehensive Benchmark for Video Anomaly Detection in Smart Homes Using Multi-Modal Large Language Models}

\author{
  Xinyi Zhao$^{1 *}$,\,
  Congjing Zhang$^{1 *}$,\,
  Pei Guo$^2$,\,
  Wei Li$^2$,\,
  Lin Chen$^{2 \dagger}$,\,
  Chaoyue Zhao$^1$,\,
  Shuai Huang$^1$ \\[6pt]
  \textsuperscript{1}University of Washington \quad
  \textsuperscript{2}Wyze Labs, Inc. \\
  {\tt\small \{xyzhao24, congjing\}@uw.edu, \{pguo, wei.li, lchen\}@wyze.com, \{cyzhao, shuaih\}@uw.edu}}

\begin{document}

\definecolor{cvprblue}{rgb}{0.21,0.49,0.74}

\maketitle

\begingroup
\renewcommand\thefootnote{*}
\footnotetext{Equal contribution. Work done during the authors’ internship at Wyze.}
\endgroup

\begingroup
\renewcommand\thefootnote{$\dagger$}
\footnotetext{Corresponding Author.}
\endgroup

\input{sec/0_abstract}    
\input{sec/1_intro}

\input{sec/2_formatting}

{
    \small
    \bibliographystyle{ieeenat_fullname}

\input{main.bbl}
}


\clearpage

\input{sec/appendix}
\end{document}

%% file: preamble.tex
\usepackage{array} 
\usepackage{multirow} 
\usepackage{makecell}
\usepackage[accsupp]{axessibility}


%% file: sec/0_abstract.tex
\begin{abstract}
Video anomaly detection (VAD) is essential for enhancing safety and security by identifying unusual events across different environments. Existing VAD benchmarks, however, are primarily designed for general-purpose scenarios, neglecting the specific characteristics of smart home applications. 
To bridge this gap, we introduce SmartHome-Bench, the first comprehensive benchmark specially designed for evaluating VAD in smart home scenarios, focusing on the capabilities of multi-modal large language models (MLLMs).
Our newly proposed benchmark consists of 1,203 videos recorded by smart home cameras, organized according to a novel anomaly taxonomy that includes seven categories, such as Wildlife, Senior Care, and Baby Monitoring. Each video is meticulously annotated with anomaly tags, detailed descriptions, and reasoning.
We further investigate adaptation methods for MLLMs in VAD, assessing state-of-the-art closed-source and open-source models with various prompting techniques. Results reveal significant limitations in current models' ability to detect video anomalies accurately.
To address these limitations, we introduce the Taxonomy-Driven Reflective LLM Chain (TRLC), a new LLM chaining framework that achieves a notable 11.62\% improvement in detection accuracy. The benchmark dataset and code are publicly available at \url{https://github.com/Xinyi-0724/SmartHome-Bench-LLM}.

\end{abstract}

%% file: sec/1_intro.tex
\section{Introduction}
\label{sec:intro}

\begin{figure}[t]
\centering
\begin{subfigure}[b]{0.5\textwidth}
    \centering
    \includegraphics[width=\linewidth]{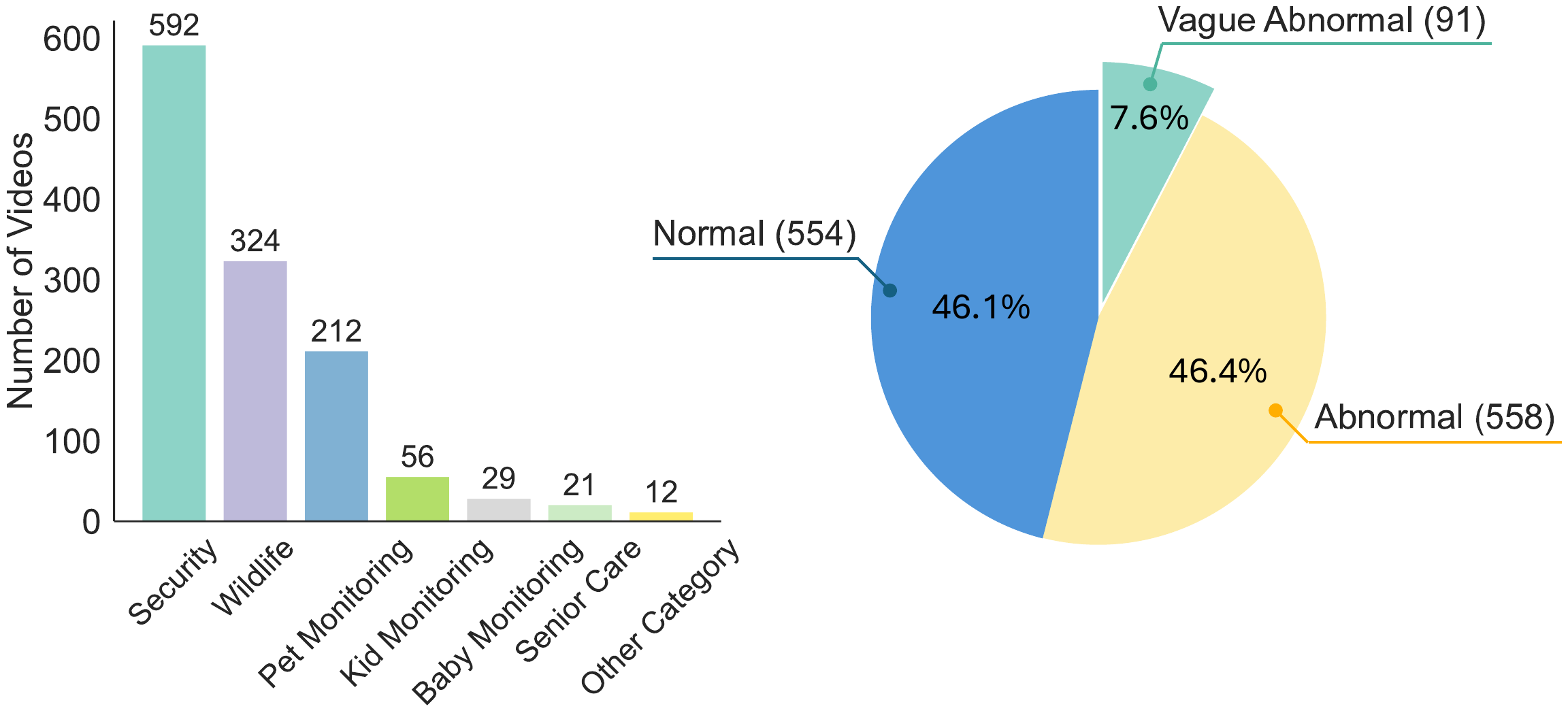}
    \caption{}
    \label{fig:video_category_dis}
\end{subfigure}

\vspace{1em}

\begin{subfigure}[b]{0.4\textwidth}
    \centering
    \includegraphics[width=\linewidth]{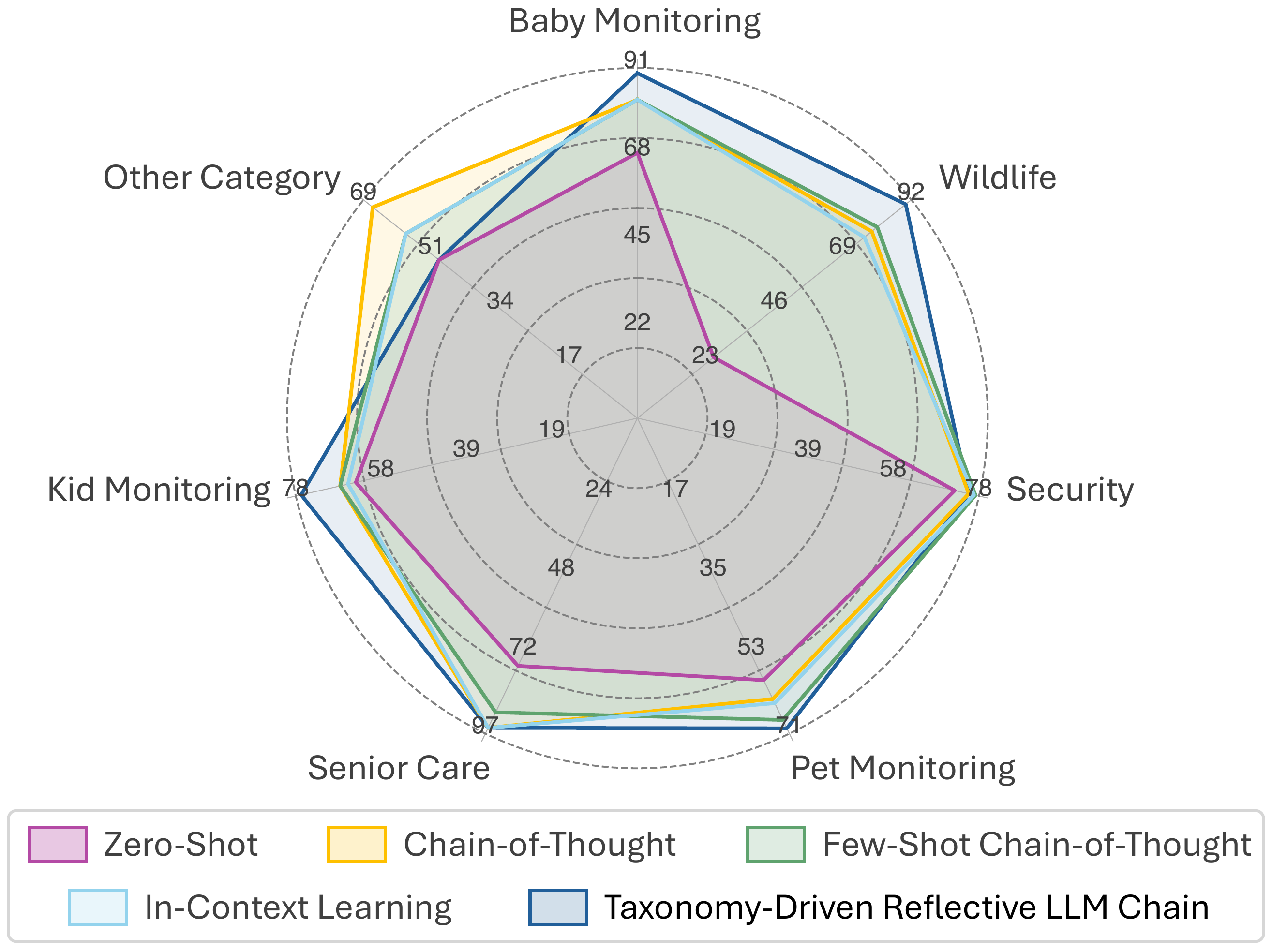}
    \caption{}
    \label{fig:rador}
\end{subfigure}

\caption{(a) Statistics for event categories and anomaly tags in the SmartHome-Bench dataset. (b) Overall anomaly detection accuracy of various adaptation methods across seven event categories, using Gemini-1.5-pro.}
\label{fig:first_overview}
\vspace{-1em}
\end{figure}

Video anomaly detection (VAD) identifies unexpected events to monitor and mitigate risks, thus improving security across diverse public spaces, including campuses, pedestrian zones, and crowded scenes \cite{sultani2018real, hasan2016learning, gong2019memorizing,ionescu2019object,pandya2018smart}. A range of supervised, weakly-supervised, one-class classification, and unsupervised methods has been proposed to generate anomaly scores for videos \cite{li2022self, tian2021weakly, wu2021learning, lv2021learning, zaheer2022generative}. However, most of these methods cannot provide descriptive rationales to support their predictions. Offering clear rationales can help users understand which behaviors or events are flagged as anomalies and why, fostering trust in the system’s assessments. Multi-modal large language models (MLLMs), with their substantial model size and capability to learn from extensive training data \cite{team2023gemini, alayrac2022flamingo, driess2023palm, li2023blip, li2023videochat}, demonstrate exceptional performance in multimodal tasks. Additionally, their generative nature enables them to make anomaly predictions and generate rationales, improving the transparency and trustworthiness of VAD~\cite{oelschlager2024evaluating, nam2024using}. 

Researchers have assessed MLLMs for VAD in various domains~\cite{bharadwaj2024vane, xu2024customizing, zhang2024holmes, lv2024video}. For example, LAVAD \cite{zanella2024harnessing} focused on detecting crimes and violent behaviors using the UCF-Crime \cite{sultani2018real} and XD-Violence \cite{wu2020not} datasets, while AnomalyRuler \cite{yang2024follow} focused on pedestrian anomalies related to biking or jumping using the ShanghaiTech \cite{liu2018future}, UCSD Ped2 \cite{li2013anomaly}, and CUHK Avenue \cite{lu2013abnormal} datasets. However, these studies focus on public spaces, overlooking anomalies within private environments like smart home scenarios. Unlike the goals of VAD in public environments, VAD in smart homes centers on more personal concerns, such as minimizing property damage, protecting vulnerable residents (e.g., young children and elderly family members), and monitoring pets and wildlife \cite{ali2023real, ren2021deep, zhu2021video}. While anomalies in smart homes may overlap with incidents in public spaces, such as crimes, they also involve many unique events rarely seen in public, like a baby climbing out of a crib or a bear entering a backyard. It remains unclear whether existing methods can effectively handle VAD in smart home scenarios. This study aims to fill the gap by evaluating the feasibility of MLLMs for VAD in smart home scenarios. 

In particular, we identify two major research gaps: (1) the absence of a dedicated benchmark for VAD in smart home scenarios, and (2) the under-exploration of adaptation strategies for MLLMs in VAD. To address the first gap, we propose \textbf{SmartHome-Bench}, a benchmark dataset of 1,203 videos featuring distinct anomaly events, such as wildlife encounters, senior care incidents, and baby monitoring issues, all collected from smart home cameras. Each video is manually annotated with anomaly tags,  detailed descriptions, and reasoning, positioning SmartHome-Bench as an ideal instructional dataset for advancing MLLM research and development in VAD. Dataset statistics are provided in Figure~\ref{fig:video_category_dis}.

To address the second gap, we conduct experiments focused on two key aspects: adaptation methods and base MLLMs. We implemented a diverse set of adaptation techniques for MLLMs, including standard prompting (zero-shot, chain-of-thought, and few-shot), contextual strategies (in-context learning), and our proposed Taxonomy-Driven Reflective LLM Chain (TRLC). These adaptations are applied across both state-of-the-art open-source and proprietary MLLMs. By evaluating these off-the-shelf models, we aim to harness their instruction-following capabilities, assessing both their anomaly detection performance and the quality of model-generated descriptions and rationales. 

Our findings indicate that current MLLMs often struggle to deliver satisfactory performance using basic prompting alone. In contrast, the TRLC framework, which integrates taxonomy-driven rules and self-reflection modules into MLLM chains, significantly enhances MLLM capabilities for VAD in smart home scenarios. This method achieves a remarkable 11.62\% improvement in anomaly detection accuracy over zero-shot prompting and outperforms all standalone prompting approaches across five out of seven event categories, as shown in Figure~\ref{fig:rador}.

In summary, our contributions are threefold:
\begin{itemize}
  \item We introduce SmartHome-Bench, the first benchmark for VAD in smart home scenarios, featuring a dataset of 1,203 videos annotated across seven event categories. 
  \item We evaluate both closed-source and open-source MLLMs using various adaptation methods, offering insights for optimizing model performance and prompt design.
  \item We propose the TRLC, a novel LLM chaining framework that improves overall VAD accuracy by 11.62\% compared to the zero-shot prompt approach.
\end{itemize}

%% file: sec/2_formatting.tex
\section{Related Work}
\label{sec:formatting}
\paragraph{Video Anomaly Detection.} MLLMs have been extensively applied in VAD recently. For instance, Holmes-VAD \cite{zhang2024holmes} processes untrimmed video with user prompts to produce frame-level anomaly scores and explanations for detected anomalies. CALLM \cite{ntelopoulos2024callm} integrates a 3D autoencoder and a visual language model into a cascade system to predict anomalies. However, MLLMs have rarely been tested in VAD for smart home scenarios, where most methods primarily rely on motion detection algorithms, statistical models, or basic machine learning techniques to detect unusual behaviors or patterns \cite{ren2021deep, yahaya2021towards, markovitz2020graph}. For example, Withanage et al. \cite{withanage2016fall} used depth cuboid similarity features with RGB-D imaging to detect falls, aiming to support in-situ assistance for fall incidents in the context of independent living for the elderly. Liu et al. \cite{liu2021privacy} transformed fall detection into a sparse recognition problem of the signal, incorporating visual shielding for enhanced privacy protection and recognition accuracy. Despite the potential of MLLMs, there remains a lack of benchmark datasets for smart home scenarios, preventing comprehensive evaluation and adaptation of these models. Our work addresses this gap by introducing SmartHome-Bench, a benchmark specifically designed for VAD in smart home scenarios.
\vspace{-1em}
\paragraph{Benchmark for MLLMs.} Recent advancements in MLLMs \cite{team2024gemini, alayrac2022flamingo, driess2023palm, li2023blip, achiam2023gpt, lin2024vila} have opened new avenues for processing diverse data types, including video, audio, and text. As a result, benchmarks designed to assess MLLM performance on video-related tasks have become increasingly important. Existing benchmarks like Flamingo \cite{alayrac2022flamingo} and VideoVista \cite{li2024videovista} demonstrate the effectiveness of MLLMs in video understanding and reasoning for fine-grained video tasks across broad domains. To explore specific task capabilities, benchmarks such as MVBench \cite{li2024mvbench} and NExT-QA \cite{xiao2021next} evaluate temporal understanding in visual language models for temporally-sensitive videos, while Video-ChatGPT \cite{maaz2023video} quantifies video dialogue capabilities for benchmarking video conversation models. VANE-Bench \cite{bharadwaj2024vane} uses question-answer pairs to evaluate VAD on both real-world and AI-generated videos. Other benchmarks, such as Video-MME \cite{fu2024video} and TempCompass \cite{liu2024tempcompass}, focus on categorizing video datasets for specific evaluation needs, like trending topics on YouTube (Video-MME \cite{fu2024video}) or temporal aspects (TempCompass \cite{liu2024tempcompass}). However, these benchmarks primarily address general video domains and overlook the unique characteristics of smart home scenarios. In contrast, SmartHome-Bench is the first benchmark specifically tailored for smart home scenarios, offering a dataset with detailed video descriptions and reasoning for detected anomalies.

\section{SmartHome-Bench Dataset}
This section presents the raw video collection and annotation process for SmartHome-Bench, with an emphasis on the proposed taxonomy used to categorize video anomalies in smart home scenarios.

\subsection{Video Collection}
We crawl videos from public sources, such as YouTube, to create SmartHome-Bench.  To identify keywords associated with common anomalies, we review the literature on home security \cite{corona2021meva}, family care \cite{zhang2015isee}, and pet monitoring \cite{kim2022dog}, creating an initial keyword set that was refined by smart home experts. Additionally, we develop a separate keyword set to capture typical, non-anomalous events in smart homes. Using these keywords, we identify 8,611 videos on YouTube. After manual filtering, we finalize a set of 1,203 videos captured by both indoor and outdoor smart home cameras. Details on the collection and filtering process are provided in Appendix~\ref{appx:video_collection}.

\begin{figure}[t]
\centering
\includegraphics[width=0.5\textwidth]{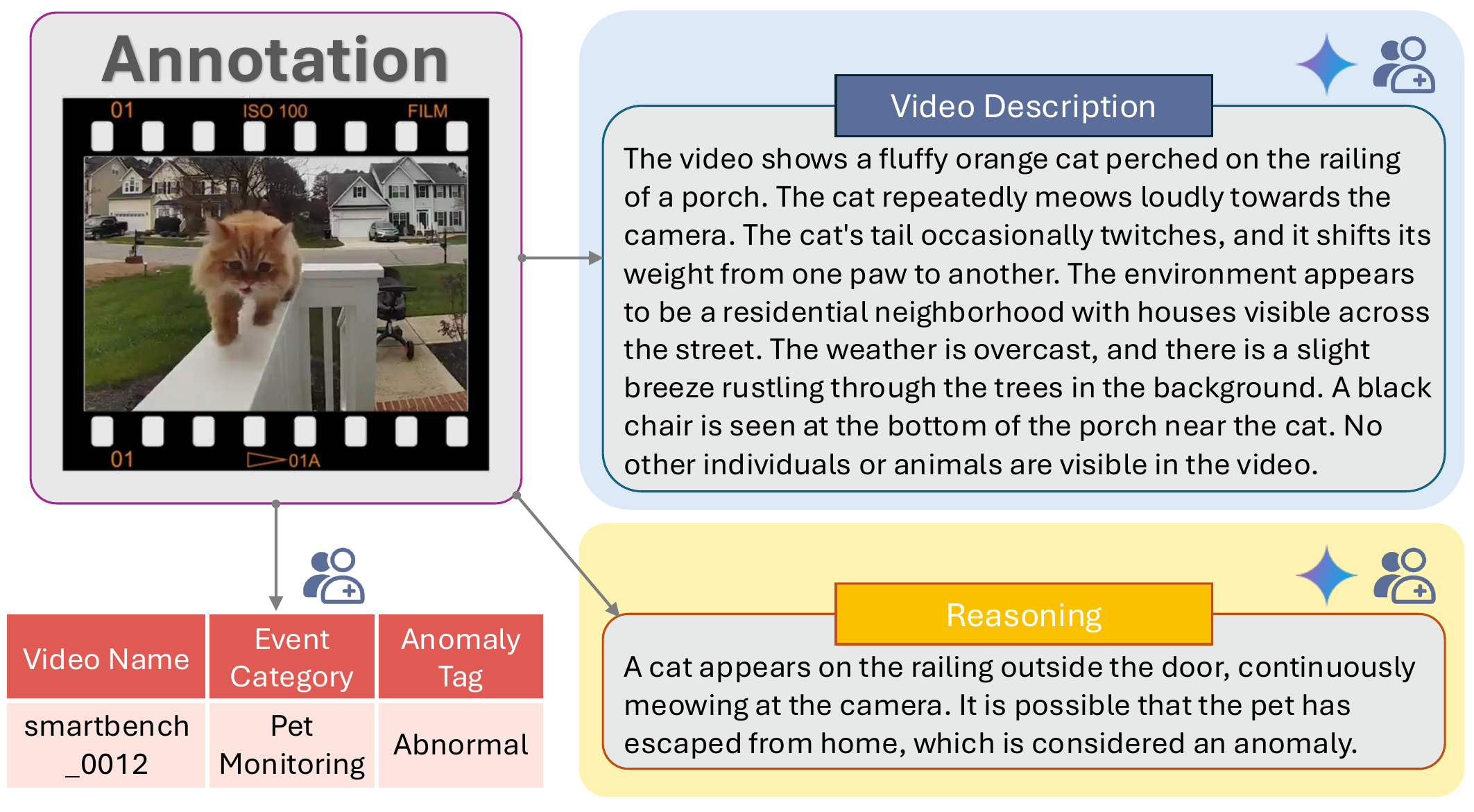}
\caption{Example of video annotation from the SmartHome-Bench dataset.}\label{fig:annotation}
\vspace{-1em}
\end{figure}

\begin{figure}[t]
\centering
\includegraphics[width=0.4\textwidth]{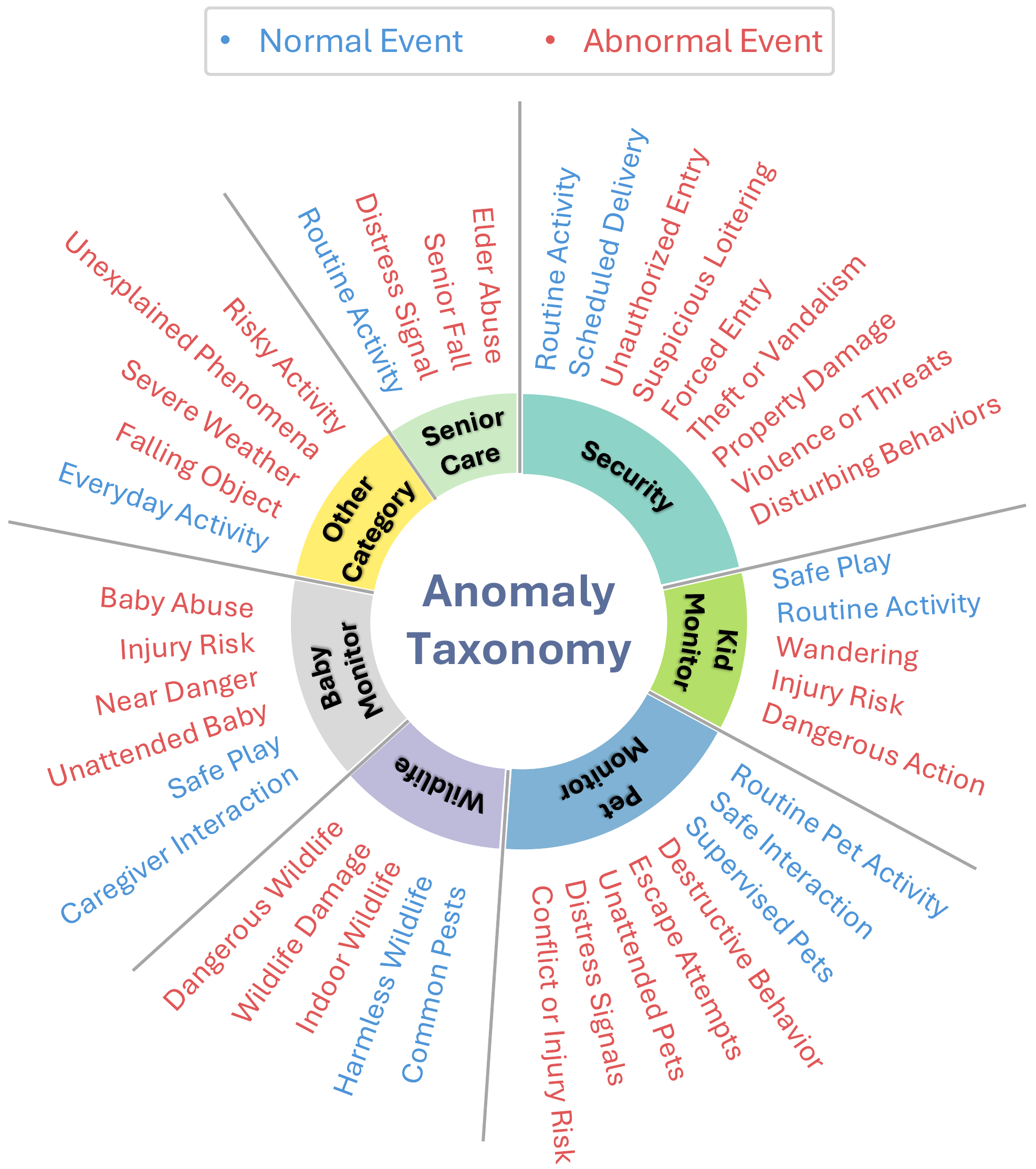}
\caption{Overview of the video anomaly taxonomy.}\label{fig:taxonomy}
\vspace{-1em}
\end{figure}


\begin{figure*}[t]
\centering
\includegraphics[width=\textwidth]{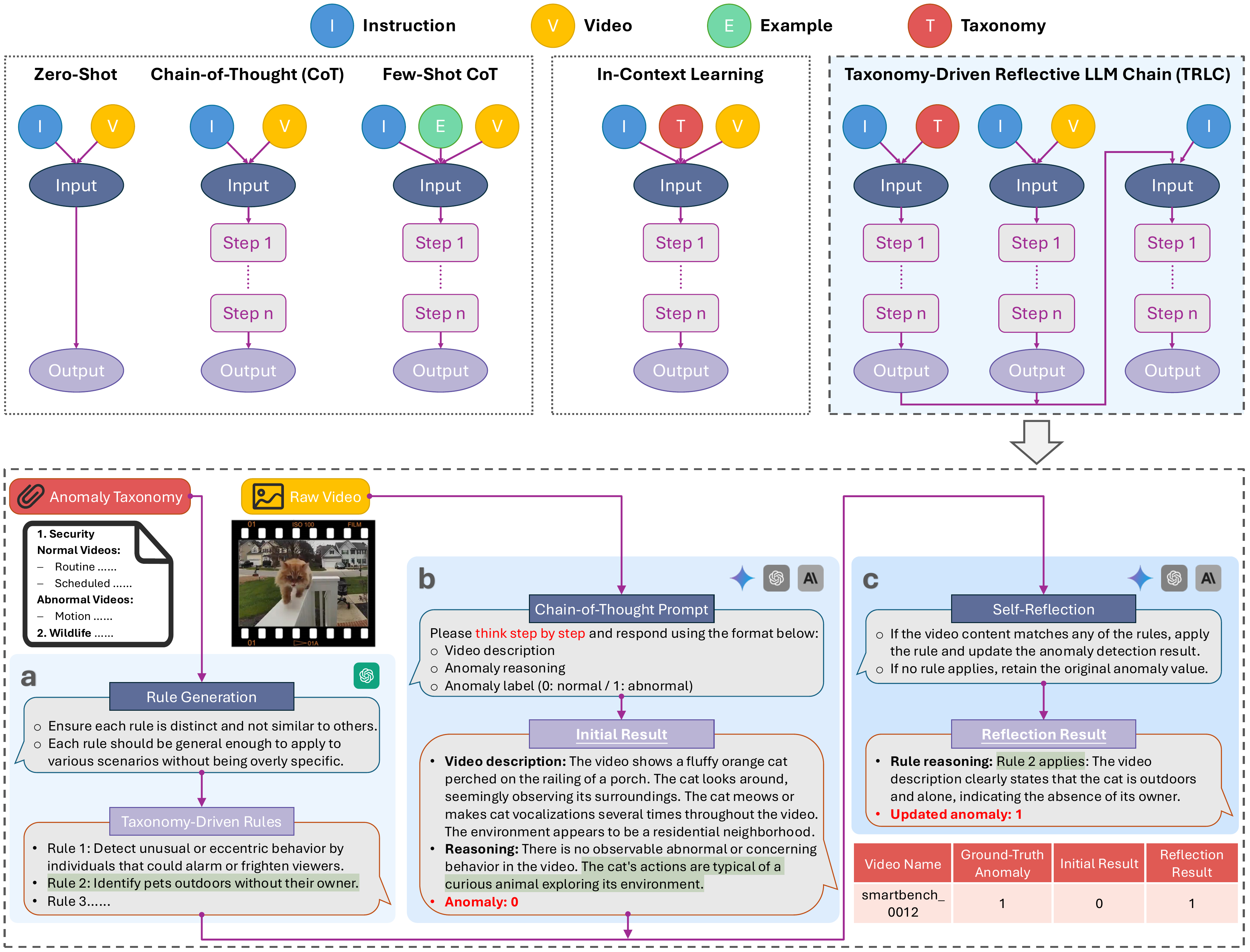}
\caption{Overview of adaptation methods and TRLC pipeline: The upper section shows vanilla adaptations, ICL methods, and the TRLC; The lower section presents the TRLC output from Gemini-1.5-pro on a SmartHome-Bench video.}\label{fig:llmchain}
\vspace{-1em}
\end{figure*}

\subsection{Video Annotation}\label{sec:annotation}
In SmartHome-Bench, each video is manually annotated with (1) the event category; (2) the anomaly tag indicating whether the video event is \texttt{normal}, \texttt{abnormal}, or \texttt{vague abnormal}; (3) textual descriptions of the events; and (4) rationales explaining the reasoning behind the assigned anomaly tag. An example of an annotated video is shown in Figure~\ref{fig:annotation}. 

Defining anomalies is a key challenge in VAD~\cite{nayak2021comprehensive}, especially in smart home scenarios where interpretations of what constitutes an anomaly can vary widely among users. To streamline the annotation process, we develop an anomaly taxonomy to guide the labeling of event categories and anomaly tags, as illustrated in Figure~\ref{fig:taxonomy}. This taxonomy defines seven primary categories: \texttt{security}, \texttt{baby monitoring}, \texttt{kid monitoring}, \texttt{senior care}, \texttt{pet monitoring}, \texttt{wildlife}, and \texttt{other category}. Each category is further divided into specific second-level event types, covering both normal and abnormal events. For example, the \texttt{senior care} category includes one normal event, \texttt{routine activity}, and three abnormal events: \texttt{distress signal}, \texttt{senior fall}, and \texttt{elder abuse}.

The complete video anomaly taxonomy is provided in Appendix \ref{appx:taxonomy}, served as a structured guideline for annotators to ensure consistency and accuracy in labeling event categories and anomaly tags. Under the guidance of taxonomy, annotators label the video with \texttt{normal} or \texttt{abnormal} tags for well-defined scenarios. If annotators could not reach a consensus on a video's anomaly classification due to limited context, it is labeled as \texttt{vague abnormal}. The distribution of categories and anomaly tags across the dataset is shown in Figure~\ref{fig:video_category_dis}, with further details on the video annotation process available in Appendix \ref{appx:video_annotation}. 

In addition to categorizing events and tagging anomalies, human annotators provide detailed descriptions of video events and articulate the reasoning behind each anomaly judgment. Video descriptions are limited to 200 words, while reasoning explanations are all in 100 words, promoting concise and precise justification. To ensure the annotation quality, there is a human review process to avoid annotator bias. These high-quality textual annotations serve as a benchmark for validating MLLMs' video understanding and reasoning processes, as demonstrated in our case analysis in Section \ref{sec:discussion}.

\section{Methods}\label{sec:method}
For smart home scenarios, users are often interested in receiving a clear alert about whether a video contains an anomalous event \cite{yamauchi2020anomaly, bakar2015activity}. By leveraging MLLMs, we aim to go beyond anomaly detection by also generating detailed descriptions and reasoning, thereby enriching the interpretability of detection outputs. We evaluate MLLMs' performance for VAD in smart home scenarios across multiple adaptation methods. As illustrated in Figure~\ref{fig:llmchain}, we begin with vanilla adaptations, such as zero-shot prompting, chain-of-thought (CoT) prompting, and few-shot CoT prompting, to gauge MLLM’s baseline capabilities in recognizing video anomalies. Then, we further utilize an in-context learning (ICL) approach that incorporates the complete anomaly taxonomy, embedding expert knowledge to enhance MLLM anomaly understanding. Building on insights that MLLMs often struggle to follow complex instructions or capture nuanced details in a single pass, we develop the TRLC, a novel LLM chaining framework, to systematically address these challenges.

\subsection{Vanilla Adaptations}
All prompts used for the following three vanilla adaptation methods are provided in Appendix \ref{appx:vanilla_prompt}.
\vspace{-1em}
\paragraph{Zero-Shot Prompting:} In this setup, MLLM is prompted directly to return a binary anomaly label, where 0 indicates no anomaly detected and 1 indicates an anomaly detected.
\vspace{-1em}

\paragraph{CoT Prompting:} CoT prompting enhances complex reasoning by incorporating intermediate reasoning steps~\cite{wei2022chain}. In this setup, we prompt MLLMs with the task instructions, smart home anomaly definitions, and video input, guiding them to complete the task in three steps: generating video descriptions, providing reasoning, and predicting the anomaly label.
\vspace{-1em}





\paragraph{Few-Shot CoT Prompting:} To enhance MLLMs' understanding of smart home video anomalies, we add a few representative anomaly examples at the end of the CoT prompt. Each example includes a video description, anomaly reasoning, and the corresponding ground-truth anomaly label.

\subsection{In-Context Learning}
We further integrate the smart home anomaly taxonomy from Section \ref{sec:annotation} into the ICL prompts, building on a similar approach that effectively guides LLMs in conversation safety assessments using a safety risk taxonomy~\cite{inan2023llama}. Building upon the CoT prompt, we include the complete anomaly taxonomy as a reference, allowing MLLMs to justify anomalies based on the taxonomy, and utilize their own knowledge if the video does not fit any predefined taxonomy category (see prompt in Appendix \ref{appx:ICL_prompt}). This integration provides MLLMs with structured guidelines and examples of both abnormal and normal events in smart home scenarios.


\subsection{Taxonomy-Driven Reflective LLM Chain } 
LLM chaining refers to a pipeline that decomposes the task into multiple steps, each solved by a unique LLM call~\cite{wu2022ai}. In our proposed TRLC framework, the VAD task is divided into three smaller subtasks: (a) Taxonomy-Driven Rule Generation, (b) Initial Prediction, and (c) Self-Reflection (see prompts for each subtask in Appendix \ref{appx:llmchain_prompt}). An example of the process in our TRLC is illustrated in Figure~\ref{fig:llmchain}. 
\vspace{-1em}
\paragraph{Step (a): Taxonomy-Driven Rule Generation} MLLMs often struggle to follow long instructions accurately and capture all detailed information in prompts. Therefore, at the first step, we make an MLLM call to condense the full taxonomy from Section \ref{sec:annotation} into a list of concise, actionable rules. This rule set is then incorporated as expert knowledge in the subsequent prompting steps. The complete set of summarized rules is provided in Appendix \ref{appx:llmchain_prompt}.
\vspace{-1em}


\paragraph{Step (b): Initial Prediction} Using the summarized rules, input videos, and a CoT prompt, we call an MLLM to generate the initial VAD prediction, which includes a video description, reasoning, and an anomaly label. This output then serves as the input for Step (c).
\vspace{-1em}

\paragraph{Step (c): Self-Reflection}
It has been observed that with a single MLLM call often leads to misclassification of certain events due to the model's limited contextual understanding. A notable example is the misclassification of an unattended cat left alone outside as a normal event, as shown in Step (b) of Figure \ref{fig:llmchain}. The model's reasoning focuses solely on typical pet behavior, overlooking potential risks a pet may face when left alone outside, such as getting lost, encountering diseases, sustaining injuries, or facing dangerous wildlife. Adding an additional self-reflection step could help correct these types of initial misclassifications.

In Step (c), we reintroduce the generated rules from Step (a) and the results from Step (b) to the MLLM, prompting it to refine the initial predictions. For instance, an unattended outdoor pet is highlighted as a common smart home anomaly in Rule \#2. With this additional context, the model successfully applies this rule to refine the initial VAD results, correcting the original classification. 

In summary, our TRLC framework enhances MLLM's contextual understanding through taxonomy-driven rules and significantly improves reasoning abilities via self-reflection. Additionally, the TRLC framework's support for configurable video anomaly taxonomies enables broader applications, such as adapting VAD for diverse public and private environments. Furthermore, TRLC enables personalized VAD by allowing users to define tailored taxonomies that align with individual standards for anomalies.

\section{Experiments}
In this section, we present the experimental results of the adaptation methods outlined in Section \ref{sec:method} across open-source and closed-source MLLMs. We convert the video's anomaly tags to binary labels: \texttt{normal(0)}, \texttt{abnormal(1)}, and \texttt{vague abnormal(1)}. The MLLM predictions, also in binary format, are then compared against these ground-truth labels. 

\subsection{Experiment Setup}
There are two ways to perform VAD: (1) asking MLLMs if the video is \texttt{abnormal}, referred to as abnormal detection and (2) asking MLLMs if the video is \texttt{normal}, referred to as normal detection. We opt for abnormal detection because it is observed that anomaly detection prompts yield better results. This is likely because MLLMs are pre-dominantly trained on normal videos and may struggle to detect anomalies without additional instructions (see results in Appendix \ref{appx:results}).

We involve six MLLMs in our experiments, including five closed-source models: Gemini-1.5-flash-001 \cite{team2024gemini}, Gemini-1.5-pro-001 \cite{team2024gemini}, GPT-4o-2024-08-06 \cite{hurst2024gpt}, GPT-4o-mini-2024-07-18 \cite{openai2023gpt4omini}, and Claude-3.5-sonnet@20240229 \cite{anthropic2024claude3sonnet}, as well as one open-source model, VILA-13b \cite{lin2024vila}. For zero-shot, CoT, few-shot CoT, and ICL methods, we test all six models, while the TRLC is evaluated only with the five closed-source models, as VILA-13b struggles to follow long, complex instructions. Overall, these models offer a comprehensive comparison and serve as the most representative benchmarks for state-of-the-art MLLM performance in anomaly detection within smart home scenarios.

\subsection{Benchmarking on Vanilla Adaptations}

\paragraph{Zero-Shot Prompting} Table~\ref{tab:AD:0shot} presents VAD performance results under zero-shot prompting, showcasing each model's inherent understanding of smart home anomalies without additional guidance. Claude-3.5-sonnet achieves the highest accuracy, recall, and F1-score, while Gemini-1.5-pro leads in precision. The accuracy of all closed-source MLLMs is only marginally above random chance (50\%), indicating limited baseline performance. Notably, VILA-13b classifies all videos as normal, underscoring its difficulty with zero-shot VAD tasks in detecting anomalies. These low VAD performance results suggest that, without guidance, these MLLMs have limited inherent understanding of smart home anomalies or may not fully utilize their capability to detect anomalies effectively.
\begin{table}[htbp]
  \centering
  \caption{Anomaly detection performance of different
MLLMs with the zero-shot prompting (Bold values indicate the highest score for each metric; applies to all tables in this paper).}
\resizebox{\linewidth}{!}{
    \begin{tabular}{lcccc}
    \toprule
    \textbf{Model} & \textbf{Accuracy} & \textbf{Precision} & \textbf{Recall} & \textbf{F1-score} \\
    \midrule
    Gemini-1.5-flash & 58.44 & 79.22 & 31.12 & 44.69 \\
    Gemini-1.5-pro & 57.36 & \textbf{84.34} & 25.73 & 39.43 \\
    GPT-4o & 68.41 & 80.09 & 55.16 & 65.33 \\
    GPT-4o-mini & 69.91 & 76.52 & 63.79 & 69.58 \\
    Claude-3.5-sonnet & \textbf{70.82} & 69.66 & \textbf{81.36} & \textbf{75.05} \\
    VILA-13b & 46.05 & 0.00  & 0.00  & 0.00 \\
    \bottomrule
    \end{tabular}
    }
  \label{tab:AD:0shot}%
\end{table}

\vspace{-2em}
\paragraph{Chain-of-Thought Prompting}
Across all test MLLMs, CoT prompting consistently improves VAD accuracy compared to zero-shot prompting (see Table~\ref{tab:AD:COT} vs. Table~\ref{tab:AD:0shot}),  underscoring the effectiveness of more granular anomaly definitions and step-by-step guidance. Among the models, Gemini-1.5-pro achieves the highest accuracy and precision. Notably, GPT-4o-mini outperforms GPT-4o in recall, albeit with reduced precision. For all closed-source MLLMs except Claude-3.5-sonnet, the gap between precision and recall narrows, resulting in a significantly improved F1-score compared to Table~\ref{tab:AD:0shot}. VILA-13b also demonstrates substantial improvement across all metrics, highlighting the positive impact of CoT prompting on its performance.

\begin{table}[htbp]
  \centering
  \caption{Anomaly detection performance of different
MLLMs with the CoT prompting.}
\resizebox{\linewidth}{!}{
    \begin{tabular}{lcccc}
    \toprule
    \textbf{Model} & \textbf{Accuracy} & \textbf{Precision} & \textbf{Recall} & \textbf{F1-score} \\
    \midrule
    Gemini-1.5-flash & 69.58 & 74.44 & 66.41 & 70.20 \\
    Gemini-1.5-pro & \textbf{74.06} & \textbf{83.77} & 64.41 & 72.82 \\
    GPT-4o & 72.57 & 83.02 & 61.79 & 70.85 \\
    GPT-4o-mini & 68.83 & 68.07 & \textbf{79.51} & \textbf{73.35} \\
    Claude-3.5-sonnet & 71.90 & 83.44 & 59.78 & 69.66 \\
    VILA-13b & 68.41 & 68.45 & 76.89 & 72.42 \\
    \bottomrule
    \end{tabular}
    }
  \label{tab:AD:COT}%
\end{table}%
\vspace{-1em}
\paragraph{Few-Shot CoT Prompting} In the few-shot CoT setup, we extend the CoT prompt by adding three representative examples of anomaly videos. Due to MLLM's processing limitations on the number of images or videos per request, these examples are provided as text tuples. As shown in Table \ref{tab:fewshot}, Gemini-1.5-pro achieves the highest accuracy, surpassing the previous CoT best of 74.06\% and leading in precision and F1-score, while GPT-4o-mini performs best in recall. However, for models like Gemini-1.5-flash, GPT-4o, GPT-4o-mini, and VILA-13b, accuracy is slightly lower than in Table \ref{tab:AD:COT}, suggesting that few-shot CoT does not fundamentally enhance CoT performance. This may be because the three examples provided in the prompt do not fully capture the range of anomalies and may distort the MLLMs' inherent knowledge, leading to misclassification.

\begin{table}[htbp]
  \centering
  \caption{Anomaly detection performance of different MLLMs with the few-shot CoT prompting.}
  \resizebox{\linewidth}{!}{
    \begin{tabular}{lcccc}
    \toprule
    \textbf{Model} & \textbf{Accuracy} & \textbf{Precision } & \textbf{Recall} & \textbf{F1-score} \\
    \midrule
    Gemini-1.5-flash & 68.41 & 79.43 & 55.93 & 65.64 \\
    Gemini-1.5-pro & \textbf{76.39} & \textbf{86.87} & 66.26 & \textbf{75.17} \\
    GPT-4o & 71.65 & 83.19 & 59.48 & 69.36 \\
    GPT-4o-mini & 68.00 & 66.30 & \textbf{82.74} & 73.61 \\
    Claude-3.5-sonnet & 72.98 & 77.65 & 70.11 & 73.68 \\
    VILA-13b & 67.17 & 69.18 & 70.57 & 69.87 \\
    \bottomrule
    \end{tabular}}
  \label{tab:fewshot}%
\end{table}%
\vspace{-1em}
\subsection{Benchmarking on ICL}



In CoT and few-shot CoT experiments, we find that adding more informative and precise anomaly definitions to the prompt improves VAD performance. With this insight, we utilize an ICL approach that incorporates the complete anomaly taxonomy in the prompt, providing MLLMs with structured categories and anomaly definitions specific to diverse smart home scenarios. 
\begin{table}[htbp]
  \centering
  \caption{Anomaly detection performance of different MLLMs with the ICL method.}
  \resizebox{\linewidth}{!}{
    \begin{tabular}{lcccc}
    \toprule
    \textbf{Model} & \textbf{Accuracy} & \textbf{Precision } & \textbf{Recall} & \textbf{F1-score} \\
    \midrule
    Gemini-1.5-flash & 67.08 & 80.78 & 51.16 & 62.64 \\
    Gemini-1.5-pro & \textbf{74.40} & 86.20 & 62.56 & \textbf{72.50} \\
    GPT-4o & 72.65 & \textbf{89.41} & 55.93 & 68.82 \\
    GPT-4o-mini & 71.74 & 83.96 & 58.86 & 69.20 \\
    Claude-3.5-sonnet & 73.82 & 84.22 & \textbf{63.33} & 72.30 \\
    VILA-13b & 65.59 & 75.82 & 53.16 & 62.50 \\
    \bottomrule
    \end{tabular}}
  \label{tab:llama}%
\end{table}%

Table~\ref{tab:llama} shows each MLLM's ability to directly apply the anomaly taxonomy in VAD with the ICL method. While half of the models (i.e., GPT-4o, GPT-4o-mini, and Claude-3.5-sonnet) demonstrate improved accuracy, the other half do not, suggesting this approach does not consistently enhance few-shot CoT performance. Except for a slight decrease in precision for Gemini-1.5-pro, all other MLLMs show increased precision, indicating that the taxonomy helps MLLMs identify anomalies more accurately.

\subsection{Benchmarking on TRLC}
ICL experiment results indicate that directly integrating the full anomaly taxonomy does not significantly improve MLLMs' VAD performance. Additionally, lengthy prompts in a single call tend to dilute the primary task, making it challenging for MLLMs to stay focus on VAD. To address this, our TRLC approach uses anomaly-specific rules generated from the taxonomy rather than the full taxonomy, providing targeted guidance and avoiding the excess detail that can lead to confusion in ICL.

\begin{figure}[!h]
\centering
\includegraphics[width=0.42\textwidth]{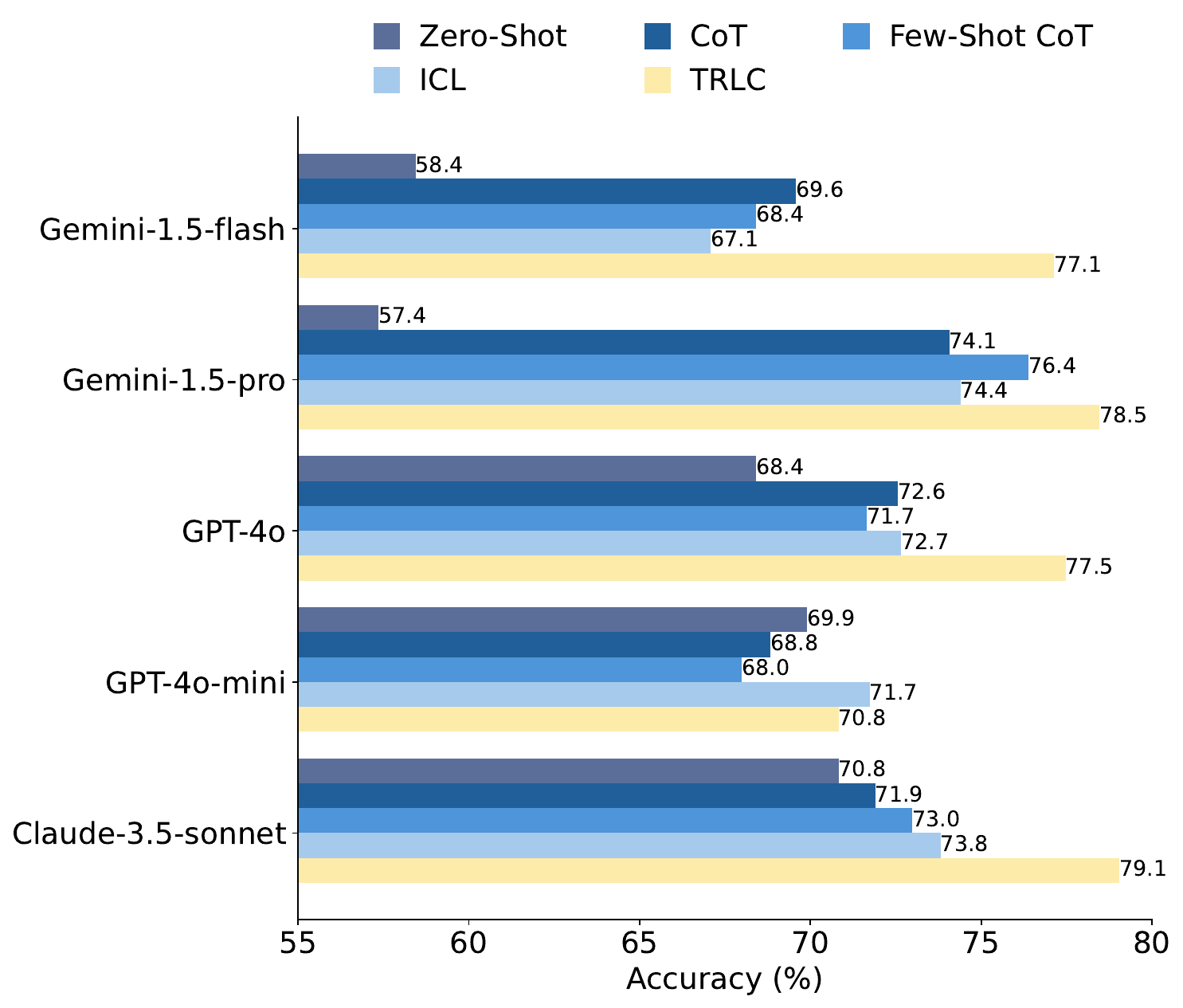}
\caption{Overall VAD accuracy of all tested adaptation methods across different MLLMs.}\label{fig:alladaptation}
\vspace{-1em}
\end{figure}

As shown in Table \ref{tab:llmchain}, applying this approach to MLLMs achieves better accuracy than all other adaptation methods in Table \ref{tab:AD:0shot}-\ref{tab:llama}, with Claude-3.5-sonnet reaching 79.05\%. Figure \ref{fig:alladaptation} further illustrates the accuracy results for all adaptation methods. Notably, our TRLC approach significantly boosts performance across all tested MLLMs, outperforming all other methods in four of the five models. The exception is GPT-4o-mini, where the TRLC ranks second, just slightly below its ICL result. On average, the TRLC method increases accuracy by 11.62\% over the zero-shot prompting across all five closed-source models. These results demonstrate that our TRLC approach provides MLLMs with an improved contextual understanding of smart home anomalies and enhances their reasoning abilities compared to no-chaining methods.


\begin{table}[!h]
  \centering
  \caption{Anomaly detection performance of different MLLMs with the TRLC method.}
  \resizebox{\linewidth}{!}{
    \begin{tabular}{lcccc}
    \toprule
    \textbf{Model} & \textbf{Accuracy} & \textbf{Precision } & \textbf{Recall} & \textbf{F1-score} \\
    \midrule
    Gemini-1.5-flash & 77.14 & 77.74 & 80.74 & 79.21 \\
    Gemini-1.5-pro & 78.47 & \textbf{82.18} & 76.73 & 79.36 \\
    GPT-4o & 77.47 & 79.35 & 78.74 & 79.04 \\
    GPT-4o-mini & 70.82 & 67.74 & \textbf{87.67} & 76.43 \\
    Claude-3.5-sonnet & \textbf{79.05} & 79.67 & 82.13 & \textbf{80.88} \\
    \bottomrule
    \end{tabular}%
  \label{tab:llmchain}}
\end{table}%

\vspace{-1em}
\paragraph{Majority Voting}
To assess the peak performance achievable with the TRLC, we combine TRLC results from the top three MLLMs: Gemini-1.5-pro, GPT-4o, and Claude-3.5-sonnet, using majority voting to determine the final anomaly prediction for each video. There are two possible voting outcomes: unanimous agreement and absolute majority. When all three MLLMs produce the same anomaly prediction, such as Gemini-1.5-pro: 0, GPT-4o: 0, and Claude-3.5-sonnet: 0, the result is classified as unanimous, and that prediction (\texttt{normal(0)}) becomes the final label. In all other cases, the majority prediction is used as the final classification.

\begin{figure}[!h]
\centering
\includegraphics[width=0.4\textwidth]{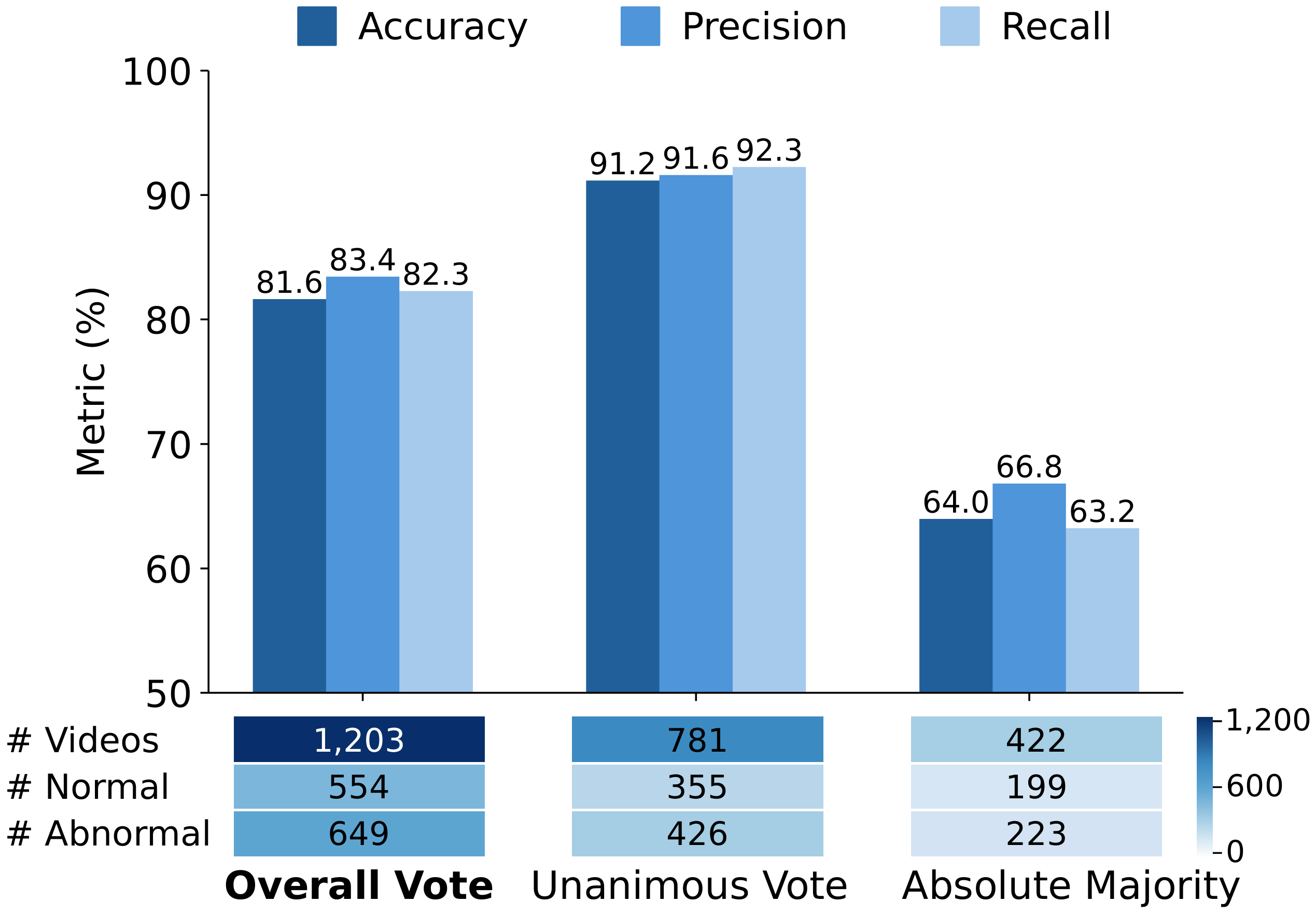}
\caption{Majority voting outcomes on VAD using TRLC results across the top three MLLMs (Gemini-1.5-pro, GPT-4o, and Claude-3.5-sonnet) with video distribution by ground-truth anomaly categories.}\label{fig:majorvote}
\vspace{-1em}
\end{figure}

As shown in Figure \ref{fig:majorvote}, this approach increases accuracy to 81.63\%, surpassing the individual performance of each model in Table \ref{tab:llmchain}. Specifically, the number of videos with unanimous agreement and absolute majority outcomes are 781 and 422, with corresponding accuracies of 91.2\% and 64.0\%, respectively. The high VAD accuracy in cases of unanimous agreement suggests potential applications, such as leveraging unanimous MLLM votes to create reliable ground-truth anomaly labels for large smart home video datasets.

\subsection{In-Depth Analysis}\label{sec:discussion}
\paragraph{Hard Case Analysis} As introduced in Section \ref{sec:annotation}, our dataset includes a category of videos with ambiguous anomalies, labeled as \texttt{vague abnormal}. These videos present challenges even for human annotators, making them a useful subset for assessing the limits of MLLMs in VAD prediction. To explore this, we analyze the accuracy of MLLMs on all 91 \texttt{vague abnormal} videos, as shown in Table \ref{tab:vague}. Generally, the accuracy for vague cases is significantly lower than the other cases across all MLLMs. Notably, with the exception of Claude-3.5-sonnet, MLLMs achieve their highest vague accuracy using the TRLC, underscoring its effectiveness in improving VAD performance, even in challenging smart home scenarios.

\begin{figure}[!h]
\centering
\includegraphics[width=0.45\textwidth]{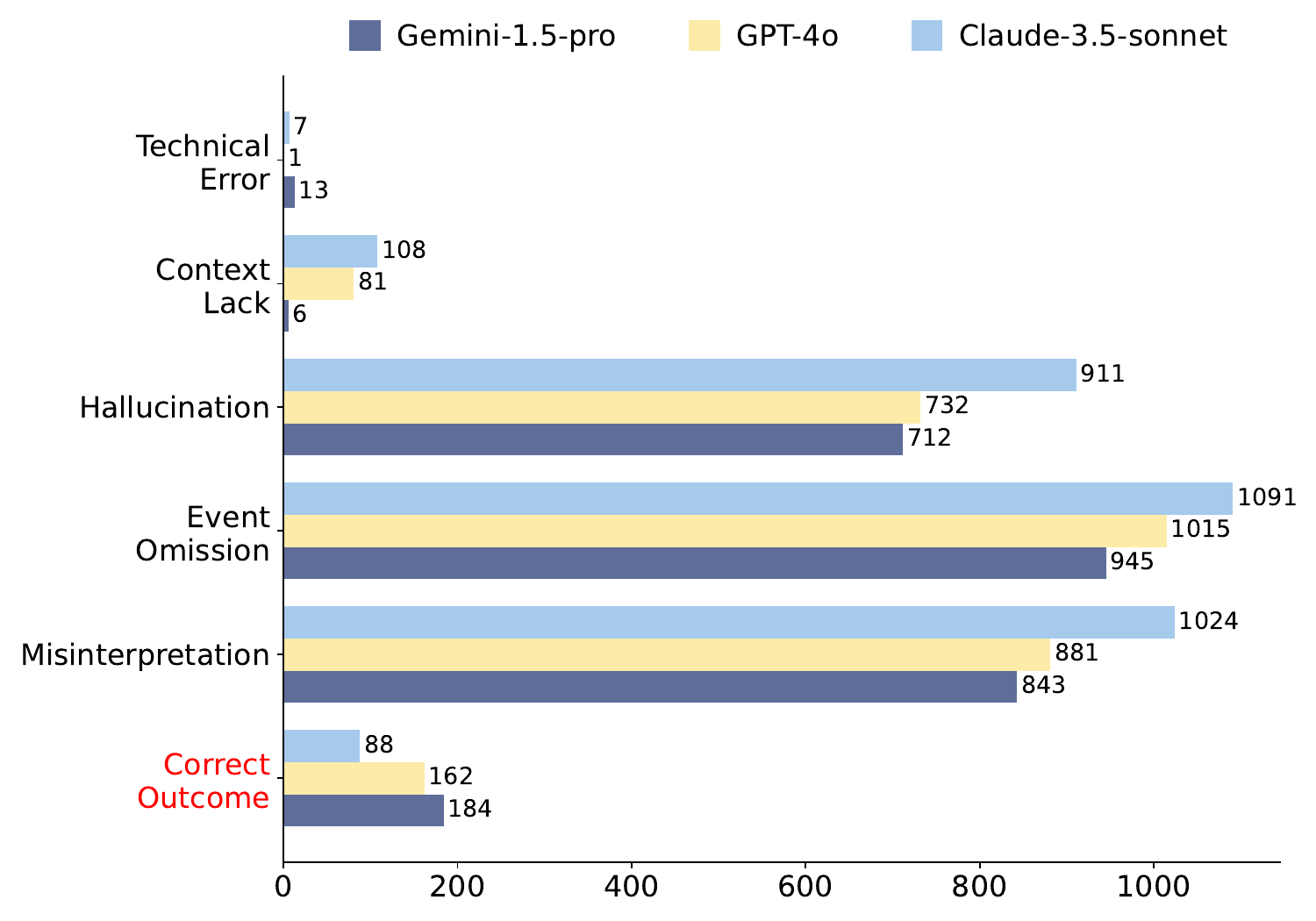}
\caption{Distribution of video outcomes for the top three MLLMs' descriptions compared to human-annotated description.}\label{fig:desdis}
\end{figure}

\begin{figure}[!h]
\centering
\includegraphics[width=0.45\textwidth]{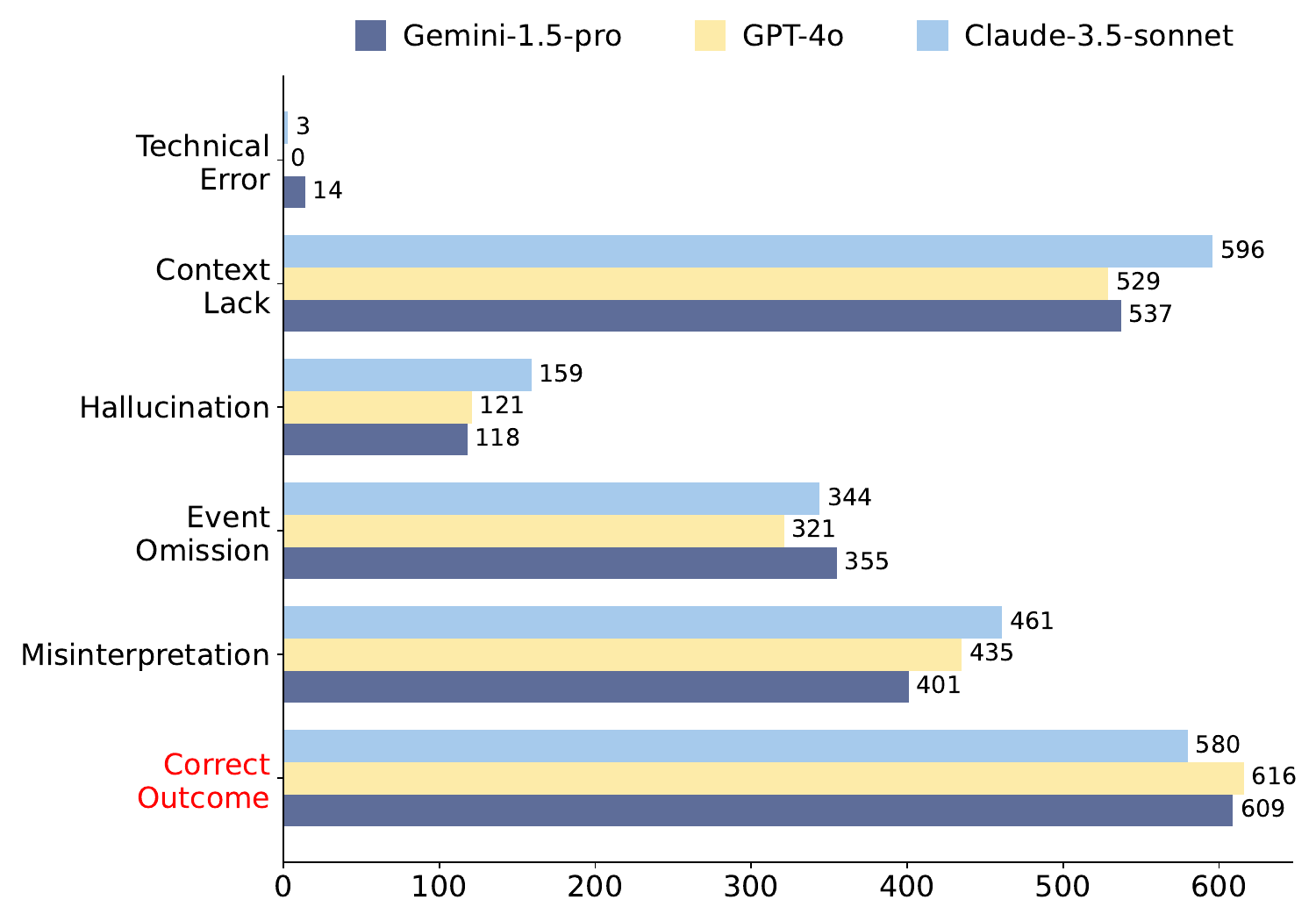}
\caption{Distribution of video outcomes for the top three MLLMs' reasoning compared to human-annotated reasoning.}\label{fig:reasondis}
\end{figure}

\begin{table}[htbp]
  \centering
  \caption{VAD accuracy on 91 vague abnormal videos across different MLLMs with all adaptation methods (ZS: zero-shot, CoT: chain-of-thought, FS: few-shot chain-of-thought, ICL: in-context learning, TRLC: taxonomy-driven reflective LLM chain). }
  \resizebox{\linewidth}{!}{
    \begin{tabular}{lccccc}
    \toprule
    \textbf{Model} & \multicolumn{5}{c}{\textbf{Accuracy}} \\
    \cmidrule(lr){2-6}
    & {\textbf{ZS}} & \textbf{CoT} & {\textbf{FS}} & \textbf{ICL} & {\textbf{TRLC}} \\
    \midrule
    Gemini-1.5-flash & 35.16 & 48.35 & 38.46 & 28.57 & \textbf{60.44} \\
    Gemini-1.5-pro & 16.48 & 37.36 & 37.36 & 35.16 & \textbf{56.04} \\
    GPT-4o & 47.25 & 37.36 & 30.77 & 23.08 & \textbf{50.55} \\
    GPT-4o-mini & 52.75 & 69.23 & 71.43 & 34.07 & \textbf{81.32} \\
    Claude-3.5-sonnet & \textbf{67.03} & 30.77 & 41.76 & 34.07 & 59.34 \\
    \bottomrule
    \end{tabular}}
  \label{tab:vague}%
\end{table}%
\vspace{-1em}
\paragraph{Error Diagnosis}
To understand which aspects MLLMs struggle with in anomaly detection within our dataset, we evaluate their video descriptions and reasoning against annotated ground truth. To capture all possible outcomes, we manually analyze MLLM outputs for 100 videos and identify five types of failure outcomes: (1) \texttt{Misinterpretation}: misdescribing or misunderstanding video events; (2) \texttt{Event Omission}: missing key abnormal events; (3) \texttt{Hallucination}: adding content that is not present; (4) \texttt{Context Lack}: failing to grasp details like the identity of the people and the emotions of the participants; and (5) \texttt{Technical Error}: failing to generate a response. Using these identified failure types, we then employ GPT-4 to evaluate the description and reasoning for all videos (see the prompts in Appendix \ref{appx:indepth_prompt}). The results are presented in Figures \ref{fig:desdis} and \ref{fig:reasondis}. Overall, MLLMs make more mistakes in video descriptions than in reasoning, likely due to the longer length of descriptions (see examples in Figure~\ref{fig:annotation}).  

Since a single video may exhibit multiple failure types, the total count of categorized types exceeds the dataset size of 1,203. Among the failure types, \texttt{Context Lack} is more prominent in reasoning than in descriptions. This occurs when MLLMs fail to grasp smart home context beyond basic descriptions, such as the identities of individuals in the video, leading to misinterpretation of normal events as anomalies or overlooking true anomalies. For instance, a description of a dog engaging with a person could have two possible interpretations: (1) playing with its owner, which is normal, or (2) attempting to fend off an intruder, which would be an anomaly, depending on whether the person is a resident. Incorporating additional context, such as a customized anomaly taxonomy and recognition of familiar faces, may help address this limitation.

\vspace{-1em}
\vspace{-1em}
\section{Conclusion}
In this paper, we introduce SmartHome-Bench, the first benchmark specifically designed for detecting anomalies in smart home scenarios. The dataset comprises 1,203 video clips, each annotated with an event category, anomaly tag, and high-quality video description and reasoning. We assess the performance of state-of-the-art closed-source and open-source MLLMs using various prompting techniques. Notably, we propose the TRLC, a novel LLM chaining framework tailored for VAD tasks, which outperforms other methods and achieves the highest accuracy of 79.05\% with Claude-3.5-sonnet.

\section{Acknowledgment}
This work is supported by Wyze Labs, Inc. and the University of Washington. We thank Kevin Beussman for donating the videos. We also thank the annotators Lina Liu, Vincent Nguyen, Pengfei Gao, Yunyun Xi, Liting Jia, and Xiaoya Hu for their hard work on data annotation.

%% file: sec/appendix.tex

\appendix
\onecolumn

\maketitle

\section{Video Collection}\label{appx:video_collection}
To curate our SmartHome-Bench dataset, we collect videos from public sources, such as YouTube. We craft a keyword set to crawl and identify videos with anomalies in smart homes. To achieve this, we survey the literature on different aspects, such as home security \cite{corona2021meva}, family care \cite{zhang2015isee}, and pet monitoring \cite{kim2022dog}. Additionally, we develop a separate keyword set to capture typical, normal events in smart homes. These keywords are then refined with input from smart home experts. Table \ref{tab:keywords} shows examples of keywords used in the search process. For each keyword, we collect approximately 20 videos from YouTube, resulting in an initial pool of 8,611 videos. We then filter out irrelevant footage, such as edited content and videos not captured by smart home cameras. For relevant videos that contain advertisements, we trim these segments to ensure the videos are clean. This curation process results in the final SmartHome-Bench dataset, comprising 1,203 videos recorded by both indoor and outdoor smart home cameras.

\begin{table*}[h!]
\centering
\caption{Example keywords for searching normal and abnormal videos.}
\begin{tabular}{| m{2.5cm} | m{13cm} |} 
\hline
\multicolumn{1}{|c|}{\textbf{Type}} & \multicolumn{1}{c|}{\textbf{Example Keywords}} \\ 
\hline
\centering Normal Videos & cat playing home cam, squirrels in the yard, rabbits outdoors, dog playtime indoors, baby sleeping crib, kid surveillance, elderly resting safe, senior camera monitoring, senior walking, visitor arrival video, vehicle arriving home, scheduled delivery home, delivery pickup, trees moving backyard, normal weather events, background motion video\\
\hline
Abnormal Videos & pet vomiting home cam, child wandering outside, kid sharp objects, child sudden fall, senior unexpected fall, senior physical distress, elderly rough caregiver, unauthorized entry attempt, package theft, car theft driveway, broken window home, suspicious person home, severe weather property, fire damage home, earthquake home safety, severe wind backyard, thunderstorm backyard, flood property risk\\
\hline
\end{tabular}
\label{tab:keywords}
\end{table*}

\section{Smart Home Anomaly Taxonomy}\label{appx:taxonomy}
We present a comprehensive taxonomy for video anomalies in the smart home domain. This taxonomy is developed based on user study, focusing on seven areas like security, senior care, and pet monitoring, and is further refined by smart home experts. Each category is further divided into normal and abnormal videos, with detailed descriptions provided for both. 

\begin{enumerate}
    \item \textbf{Wildlife}
    \begin{itemize}
        \item[] \textbf{Normal Videos:}
        \begin{itemize}
            \item \textbf{Harmless Wildlife:} Harmless wildlife sightings, such as squirrels, birds, or rabbits, moving through the yard.
            \item \textbf{Common Pests:} Common pest activity that doesn’t pose immediate danger (e.g., bugs in the garden).
        \end{itemize}
        
        \item[] \textbf{Abnormal Videos:}
            \begin{itemize}
            \item \textbf{Dangerous Wildlife:} Presence of dangerous wildlife like snakes, spiders, or raccoons that may pose a health risk.
            \item \textbf{Wildlife Damage:} Any wildlife activity that causes or potentially causes damage to property or threatens human or pet safety.
            \item \textbf{Indoor Wildlife:} Any wildlife (dangerous or not) that enters a home without clear containment.
        \end{itemize}
    \end{itemize}

    \item \textbf{Pet Monitoring}
    \begin{itemize}
        \item[] \textbf{Normal Videos:}
        \begin{itemize}
            \item \textbf{Routine Pet Activity:} Pets engaging in regular play, resting or moving around within designated safe areas.
            \item \textbf{Safe Interaction:} Pets interacting with known family members or other pets.
            \item \textbf{Supervised Pets:} Pets accompanied by their guardian without interacting with property or people in harmful ways.
        \end{itemize}
        \item[] \textbf{Abnormal Videos:}
        \begin{itemize}
            \item \textbf{Unattended Pets:} Pets left outside alone for extended periods.
            \item \textbf{Escape Attempts:} Pets attempting to escape, leaving the designated area, or exhibiting behaviors indicating escape attempts.
            \item \textbf{Destructive Behavior:} Pets causing property damage by actions like chewing, scratching, or digging.
            \item \textbf{Distress Signals:} Behaviors that indicate illness or distress, like vomiting, excessive scratching, or erratic movements.
            \item \textbf{Conflict or Injury Risk:} Any interaction with others that could lead to conflict or injury.
        \end{itemize}
    \end{itemize}

    \item \textbf{Baby Monitoring}
    \begin{itemize}
        \item[] \textbf{Normal Videos:}
        \begin{itemize}
            \item \textbf{Safe Play:} Baby engaging in play or sleep within safe zones or under supervision.
            \item \textbf{Caregiver Interaction:} Harmless interactions between the baby and caregivers.
        \end{itemize}
        \item[] \textbf{Abnormal Videos:}
        \begin{itemize}
            \item \textbf{Near Danger:} Baby nearing dangerous zones (e.g., staircases, swimming pools) without adult supervision.
            \item \textbf{Unattended Baby:} Baby wandering outside a crib, stroller, or designated play area without adult presence.
            \item \textbf{Injury Risk:} Sudden, unexpected falls that may lead to injury.
            \item \textbf{Baby Abuse:} Any abusive behavior toward the baby, such as hitting, or forcing them to act against their will.
        \end{itemize}
    \end{itemize}

    \item \textbf{Kid Monitoring}
    \begin{itemize}
        \item[] \textbf{Normal Videos:}
        \begin{itemize}
            \item \textbf{Safe Play:} Kids playing or moving around indoors or outdoors within designated areas.
            \item \textbf{Routine Activities:} Regular daily activities under adult supervision.
        \end{itemize}
        \item[] \textbf{Abnormal Videos:}
        \begin{itemize}
            \item \textbf{Wandering:} Kids found wandering outdoors or in dangerous locations without adult supervision.
            \item \textbf{Dangerous Actions:} Dangerous actions indoors (e.g., playing with sharp objects, accessing restricted areas) or significant health/safety concerns (e.g., choking hazards).
            \item \textbf{Injury Risk:} Sudden, unexpected falls that may lead to injury.
        \end{itemize}
    \end{itemize}

    \item \textbf{Senior Care}
    \begin{itemize}
        \item[] \textbf{Normal Videos:}
        \begin{itemize}
            \item \textbf{Routine Activity:} Seniors engaging in routine activities like walking, resting, or interacting with caregivers or family.
        \end{itemize}
        \item[] \textbf{Abnormal Videos:}
        \begin{itemize}
            \item \textbf{Senior Falls:} Sudden, unexpected falls that may lead to injury.
            \item \textbf{Distress Signals:} Signs of distress or calls for help through hand gestures or unusual body language.
            \item \textbf{Elder Abuse:} Any abusive or rough behavior by caregivers toward seniors, including verbal and physical abuse.
        \end{itemize}
    \end{itemize}

    \item \textbf{Security}    
    \begin{itemize}
        \item[] \textbf{Normal Videos:}
        \begin{itemize}
            \item \textbf{Routine Activity:} Routine activity of homeowners, known visitors, or vehicles arriving and leaving.
            \item \textbf{Scheduled Delivery:} Scheduled package deliveries or pickups without interference.
        \end{itemize}
        \item[] \textbf{Abnormal Videos:}
        \begin{itemize}
            \item \textbf{Unauthorized Entry:} Motion or presence indicating potential break-ins, or trespassing.
            \item \textbf{Suspicious Loitering:} Loitering individuals or those wearing unusual attire that deviates from the norm.
            \item \textbf{Forced Entry:} Forced entry attempts, such as fiddling with locks, tampering with doors or windows, or trying to enter a home or vehicle through unconventional means.
            \item \textbf{Theft or Vandalism:} Unauthorized removal of packages, vehicles, or other items.
            \item \textbf{Property Damage:} Acts of property damage like graffiti, broken windows, car crashes, or other forms of vandalism.
            \item \textbf{Violence or Threats:} Actions that might cause harm, such as kidnapping, aggressive confrontations, or any threatening behavior.
            \item \textbf{Disturbing Behavior:} Unusual or eccentric behavior by individuals that could alarm or frighten viewers.
        \end{itemize}
    \end{itemize}

    \item \textbf{Other Category}   
    \begin{itemize}
        \item[] \textbf{Normal Videos:}
        \begin{itemize}
            \item \textbf{Everyday Activity:} Videos that do not fit any of the above categories but show harmless, everyday activities, such as trees waving, normal weather events, or background motion.
        \end{itemize}
        \item[] \textbf{Abnormal Videos:}
        \begin{itemize}
            \item \textbf{Severe Weather:} Severe weather conditions or natural disasters like fires, earthquakes, floods, or storms causing property damage or safety hazards.
            \item \textbf{Unexplained Phenomena:} Unexplained phenomena of inanimate objects.
            \item \textbf{Falling Objects:} Sudden, unexpected falls of inanimate objects that may cause damage or injury.
            \item \textbf{Risky activities:} Irregular activities that do not fit into other categories but may pose risks or concerns.
        \end{itemize}
    \end{itemize} 
    
\end{enumerate}

\section{Video Annotation}\label{appx:video_annotation}

During the video annotation process, we assign unique IDs to the downloaded videos to prevent annotators from being influenced by the original titles or metadata. The annotators classify each video into one or more of the seven categories in the taxonomy outlined in Appendix \ref{appx:taxonomy}, as real-world events in a single video may span multiple categories. Each video is then assigned an anomaly tag of \texttt{normal}, \texttt{abnormal}, or \texttt{vague abnormal}, based on the definitions outlined in the taxonomy. The \texttt{vague abnormal} category is created for videos where annotators cannot reach a consensus on whether the content is normal or abnormal. This category is specifically introduced to challenge the video anomaly detection (VAD) capabilities of multi-modal large language models (MLLMs) with videos that are difficult for even humans to classify. A \texttt{vague normal} category is not included, as any ambiguity regarding the presence of an anomaly is classified under \texttt{vague abnormal}. 

We instruct annotators to write high-quality video descriptions and provide detailed reasoning for the assignment of each video's anomaly tag.  These annotations establish a strong foundation for future research by enabling the generation of diverse question-answer pairs to assess the video understanding and reasoning capabilities of MLLMs. Additionally, the inclusion of ground-truth reasoning ensure a transparent inference process for classifying normal and abnormal videos, which can be leveraged to fine-tune MLLMs and improve anomaly detection accuracy in smart home scenarios. To maintain consistency and quality across video descriptions and reasoning annotations, we use the Gemini-1.5-pro model to generate initial drafts. Annotators then review each video and refine or rewrite these drafts according to three main criteria: (1) clarity and precision of language, (2) alignment of descriptions and reasoning with the video content, and (3) accuracy in identifying key elements such as objects triggering anomalies, abnormal movements, participants, and environmental conditions. 
\begin{figure}[!ht]
\centering
\includegraphics[width=0.5\textwidth]
{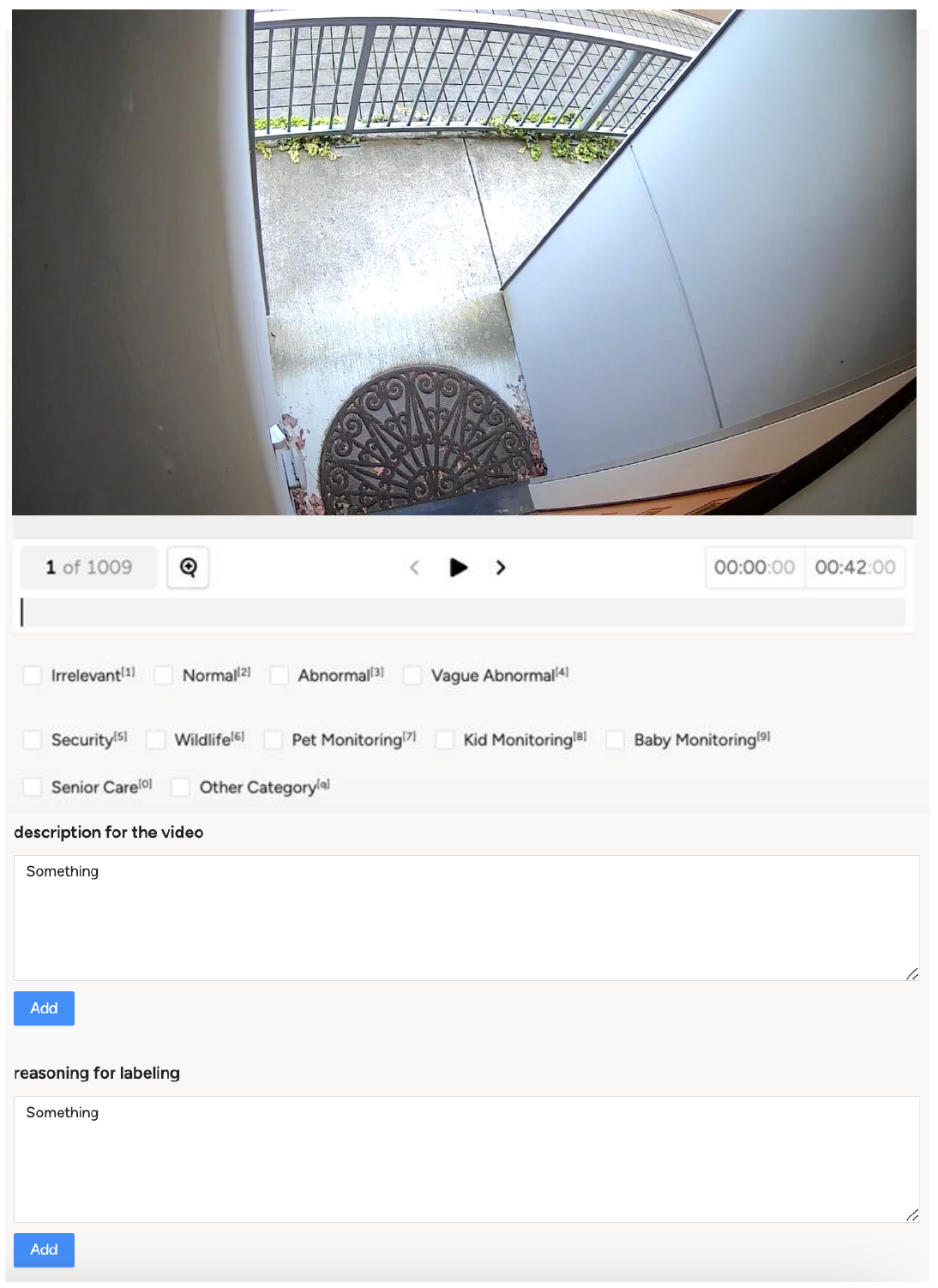}
\caption{The UI enables annotators to label videos by selecting event categories, assigning anomaly tags, and providing detailed video descriptions along with the reasoning behind the observed anomalies or normality.}\label{fig:annotation_ui}
\end{figure} 

To streamline the annotation process and maximize efficiency, annotators use a customized user interface (UI), shown in Figure~\ref{fig:annotation_ui}, to label each video's event category and anomaly tag, as well as to manually write the description and reasoning. To ensure the quality and consistency of the annotations, we conduct a human review of a randomly select 200 videos after the initial round of annotation .

\begin{figure}[!h]
\centering
\includegraphics[width=0.7\textwidth]{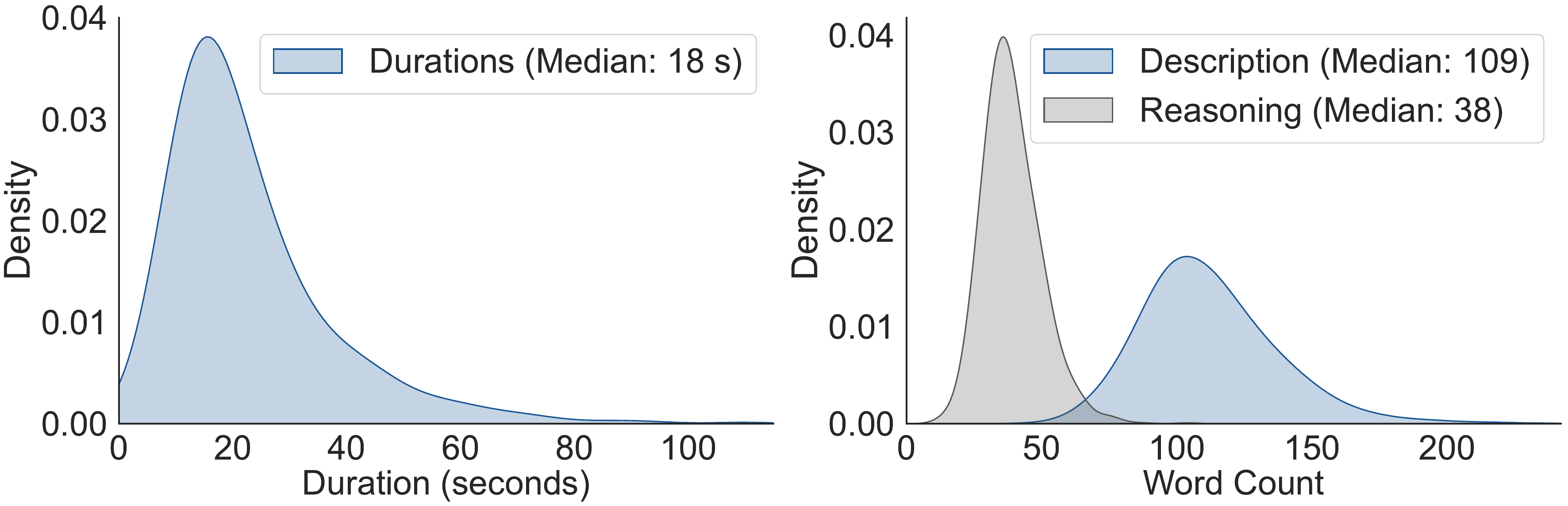}
\caption{Distribution of video durations and word counts for human-annotated video descriptions and reasoning.}\label{fig:video_time_dis}
\end{figure}

Following the annotation process for all 1,203 videos, the statistics of the SmartHome-Bench dataset are presented in Figure 1a of the main paper. The dataset shows a balanced distribution between abnormal and normal videos, with the security category containing the largest number of videos among the seven event categories. Additionally, Figure~\ref{fig:video_time_dis} illustrates the distribution of video durations and word counts for descriptions and reasoning annotations. The average video length is approximately 20 seconds, with most clips being shorter than 80 seconds. This duration aligns well with the frame-processing limitations of some existing MLLMs, enabling relatively comprehensive predictions in VAD tasks. The word count distribution reveals that reasoning annotations are typically more concise than descriptions, as they focus solely on the key event leading to the assigned anomaly tag. In contrast, descriptions provide a detailed account of all events within the video.

\newpage

\section{Prompts for Adaptation Methods and In-Depth Analysis}\label{appx:prompt}
We provide all prompts used for adaptation methods and error diagnosis in in-depth analysis as follows.

\subsection{System Prompt for Vanilla Adaptations}\label{appx:vanilla_prompt}
\begin{figure*}[!ht]
\centering
\includegraphics[width=0.6\textwidth]{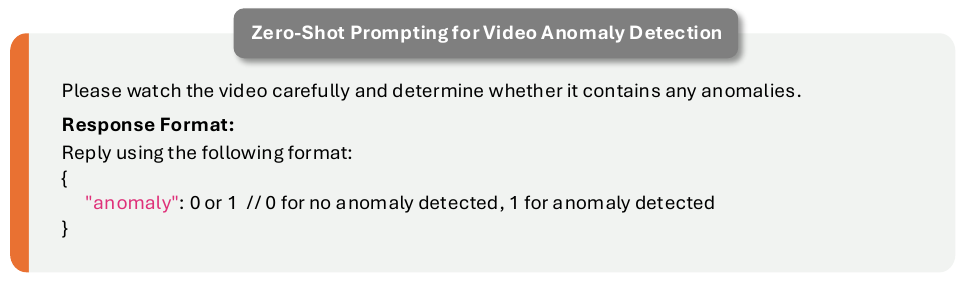}
\caption{System prompts adopted in zero-shot prompting for VAD. MLLMs are prompted directly to return a binary anomaly label.}\label{fig:0shotprompt-AD}
\end{figure*}

Figure \ref{fig:0shotprompt-AD} shows the prompts used in zero-shot prompting for the VAD task.

\begin{figure*}[!ht]
\centering
\includegraphics[width=0.7\textwidth]{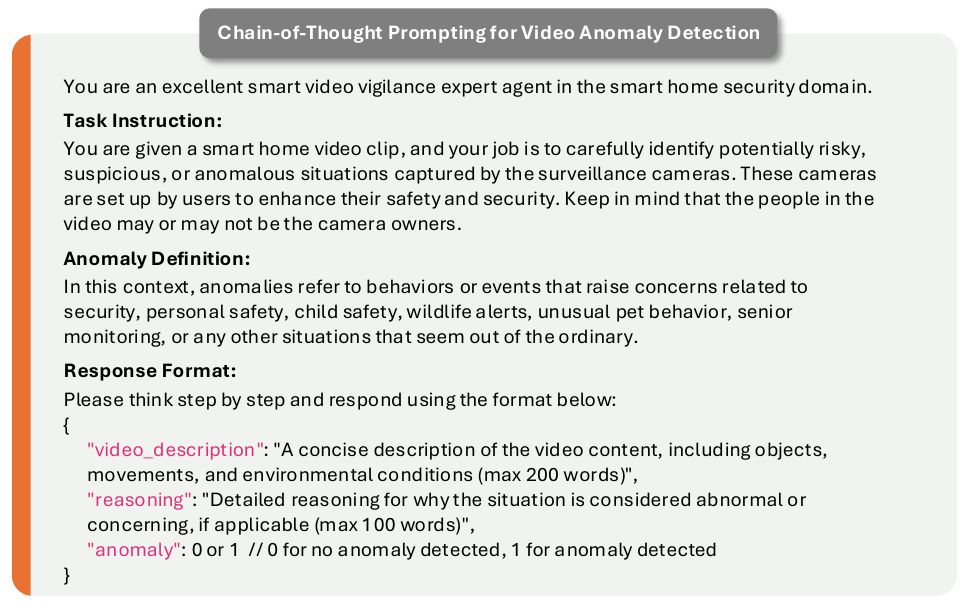}
\caption{System prompts adopted in CoT prompting include task instructions, smart home anomaly definitions, and video input, guiding MLLMs to complete the task in three steps: generating video descriptions, providing reasoning, and predicting the anomaly label.}\label{fig:cotprompt-AD}
\end{figure*}

Figure \ref{fig:cotprompt-AD} shows the prompts used in chain-of-thought (CoT) prompting for the VAD task.

\newpage

\begin{figure*}[!ht]
\centering
\includegraphics[width=0.7\textwidth]{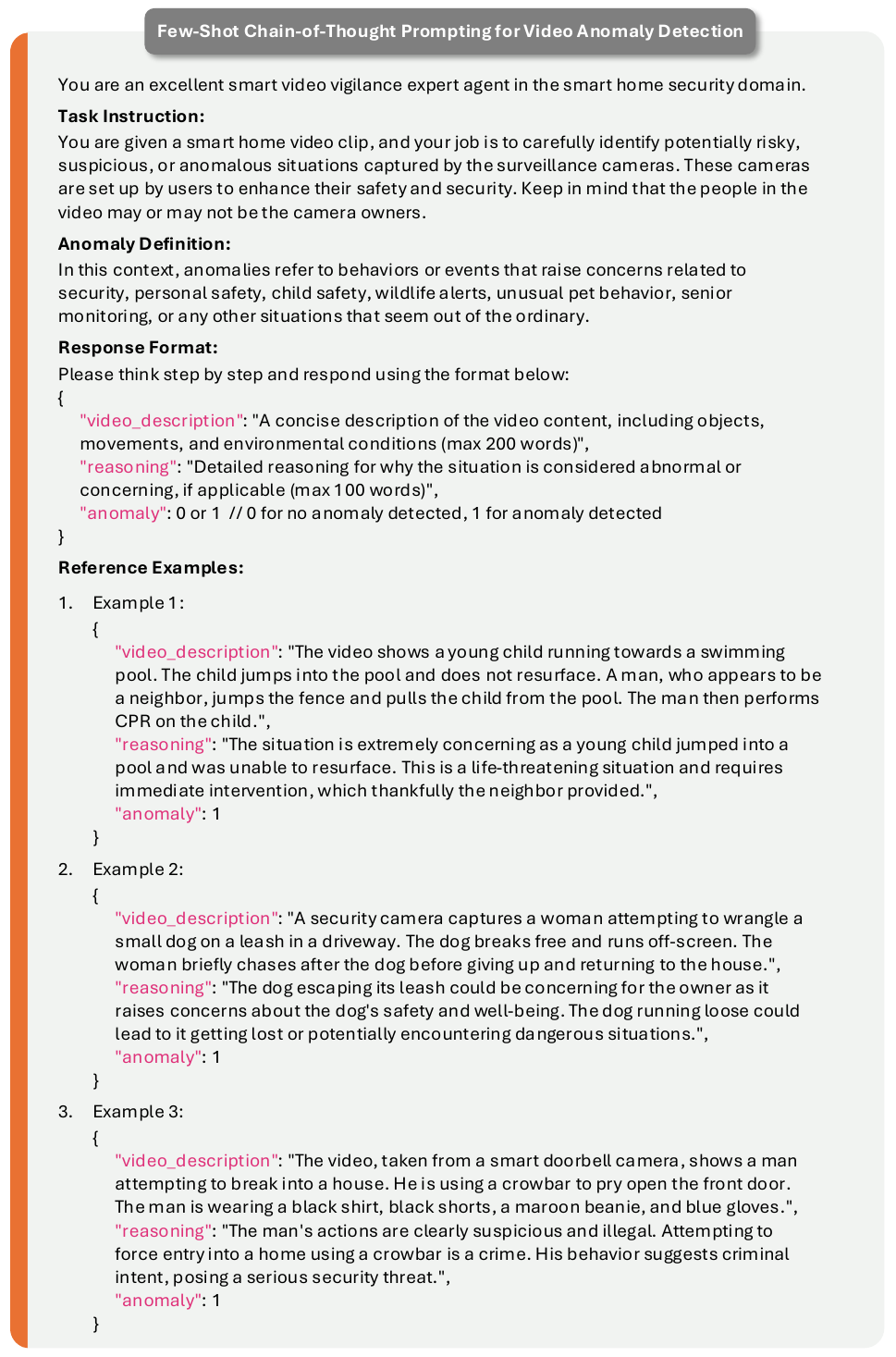}
\caption{System prompts adopted in few-shot CoT prompting for VAD. Each example provided includes a video description, anomaly reasoning, and the corresponding ground-truth anomaly label.}\label{fig:fewprompt}
\end{figure*}

Figure \ref{fig:fewprompt} shows the prompts used in the few-shot CoT prompting for the VAD task.

\newpage 

\subsection{System Prompt for In-Context Learning}\label{appx:ICL_prompt}
\begin{figure*}[!ht]
\centering
\includegraphics[width=0.7\textwidth]{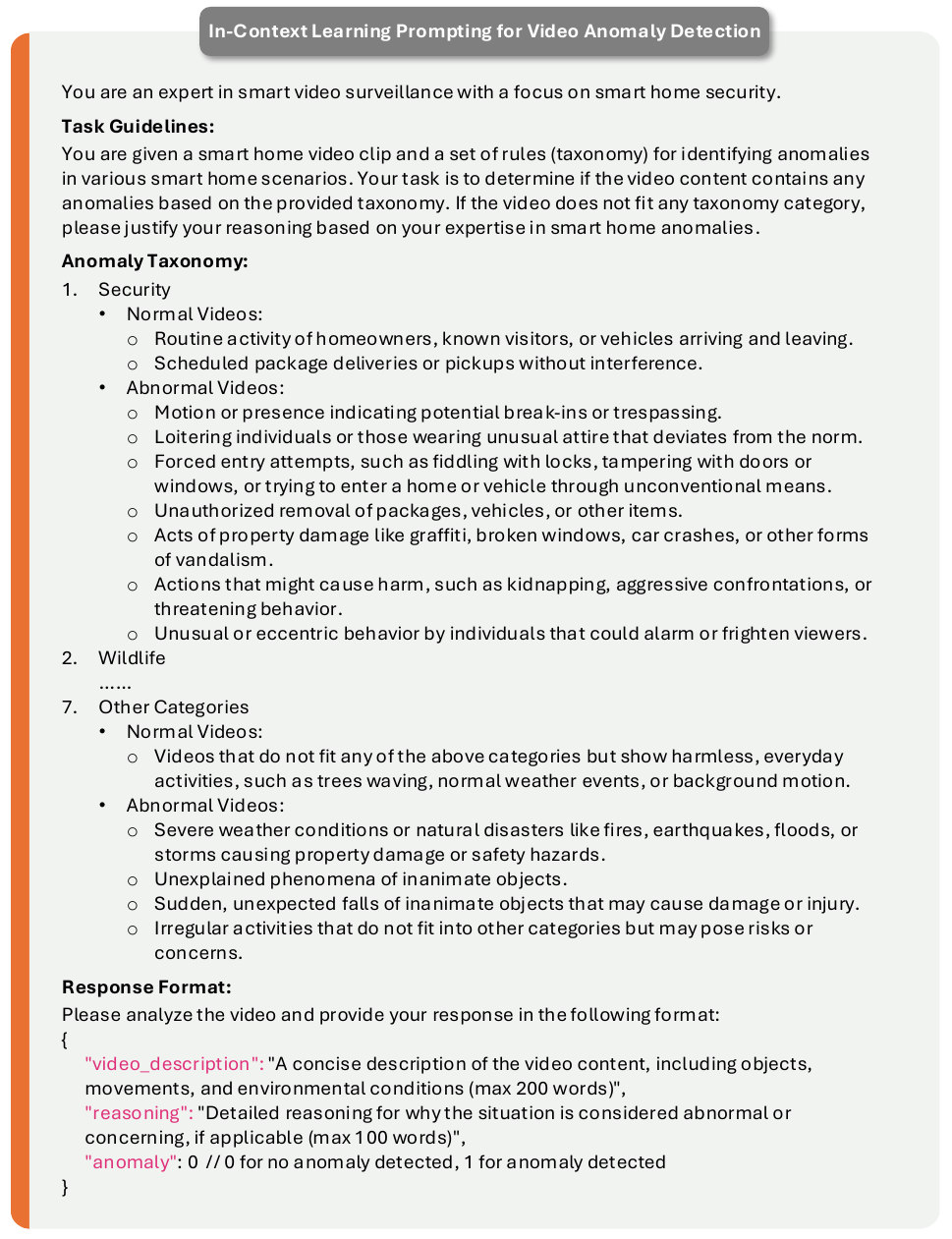}
\caption{System prompts adopted in ICL for VAD. Building upon the CoT prompt, we include the complete anomaly taxonomy as a reference.}\label{fig:llamaprompt}
\end{figure*}

The prompts used in in-context learning (ICL) for the VAD task are shown in Figure \ref{fig:llamaprompt}.

\newpage 

\subsection{System Prompt for Taxonomy-Driven Reflective LLM Chain}\label{appx:llmchain_prompt}
\begin{figure*}[!ht]
\centering
\includegraphics[width=0.68\textwidth]{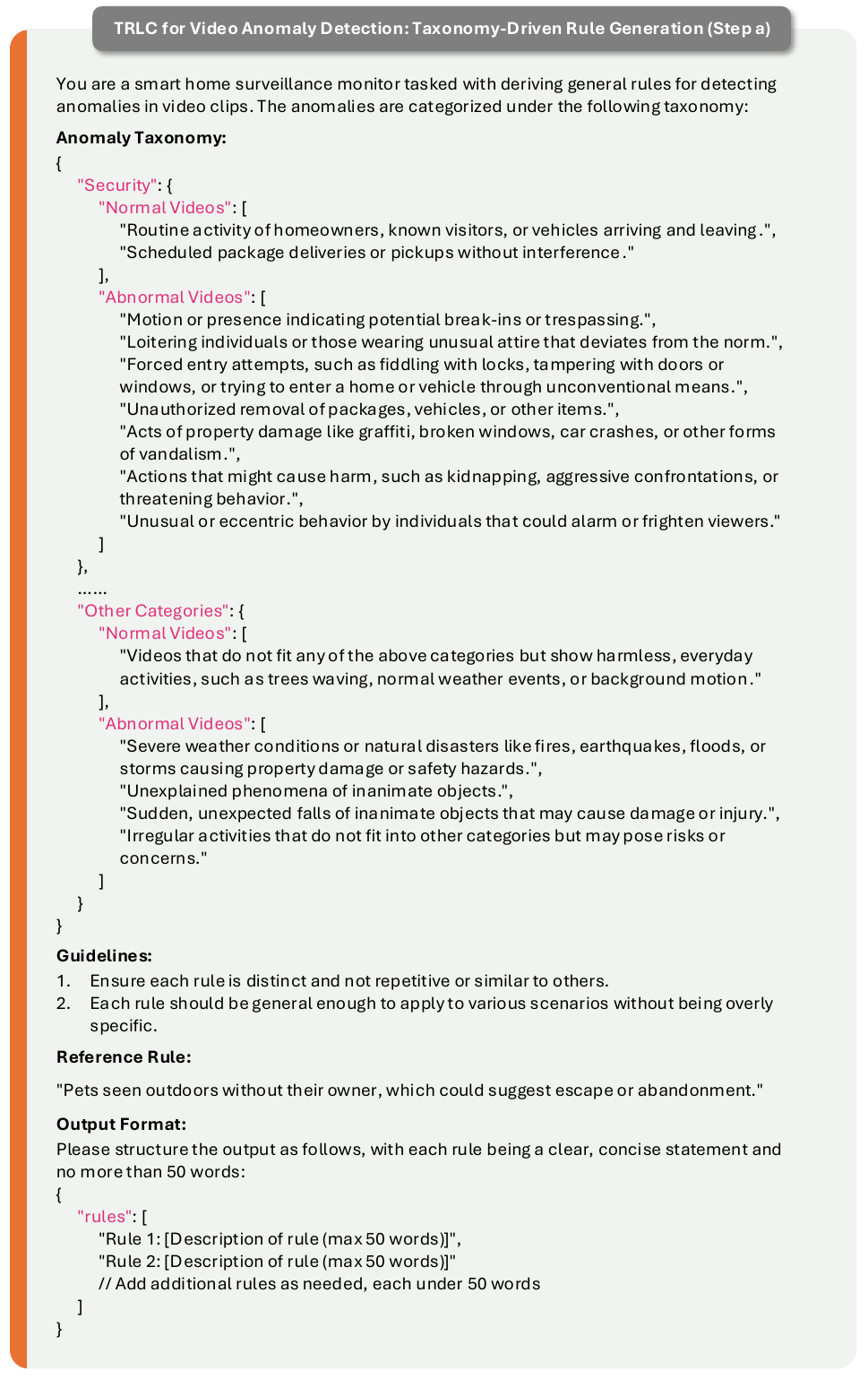}
\caption{System prompts adopted in step (a) of the TRLC for VAD: taxonomy-driven rule generation.}\label{fig:chain-a-prompt}
\end{figure*}

\begin{figure*}[!ht]
\centering
\includegraphics[width=0.7\textwidth]{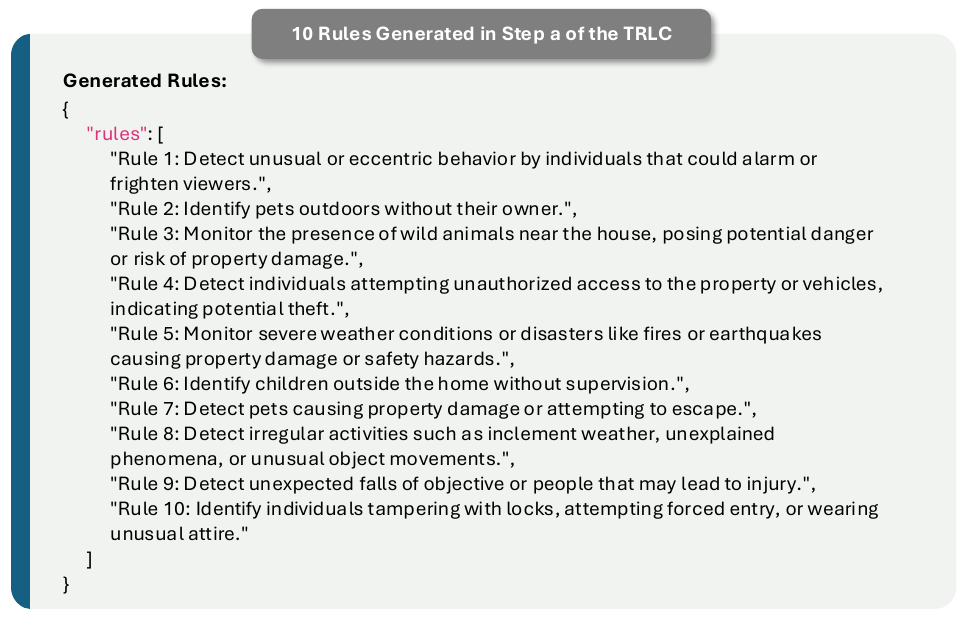}
\caption{10 rules generated from the full video anomaly taxonomy in step (a) of TRLC by GPT-4o.}\label{fig:10rule}
\end{figure*}

\begin{figure*}[!ht]
\centering
\includegraphics[width=0.7\textwidth]{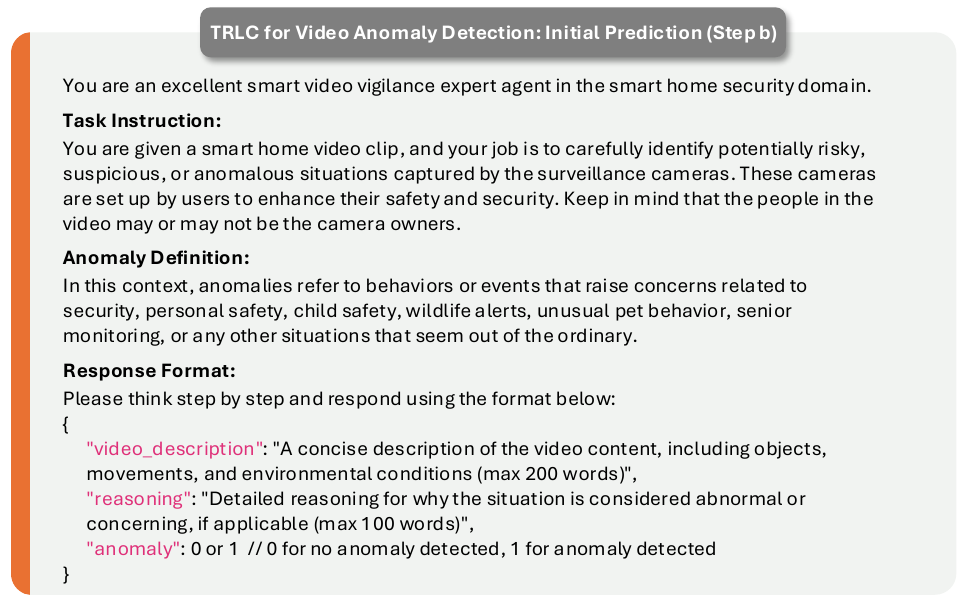}
\caption{System prompts adopted in step (b) of the TRLC for VAD: initial prediction. (These prompts are identical to the CoT prompts shown in Figure \ref{fig:cotprompt-AD}).}\label{fig:chain-b-prompt}
\end{figure*}

\begin{figure*}[!ht]
\centering
\includegraphics[width=0.7\textwidth]{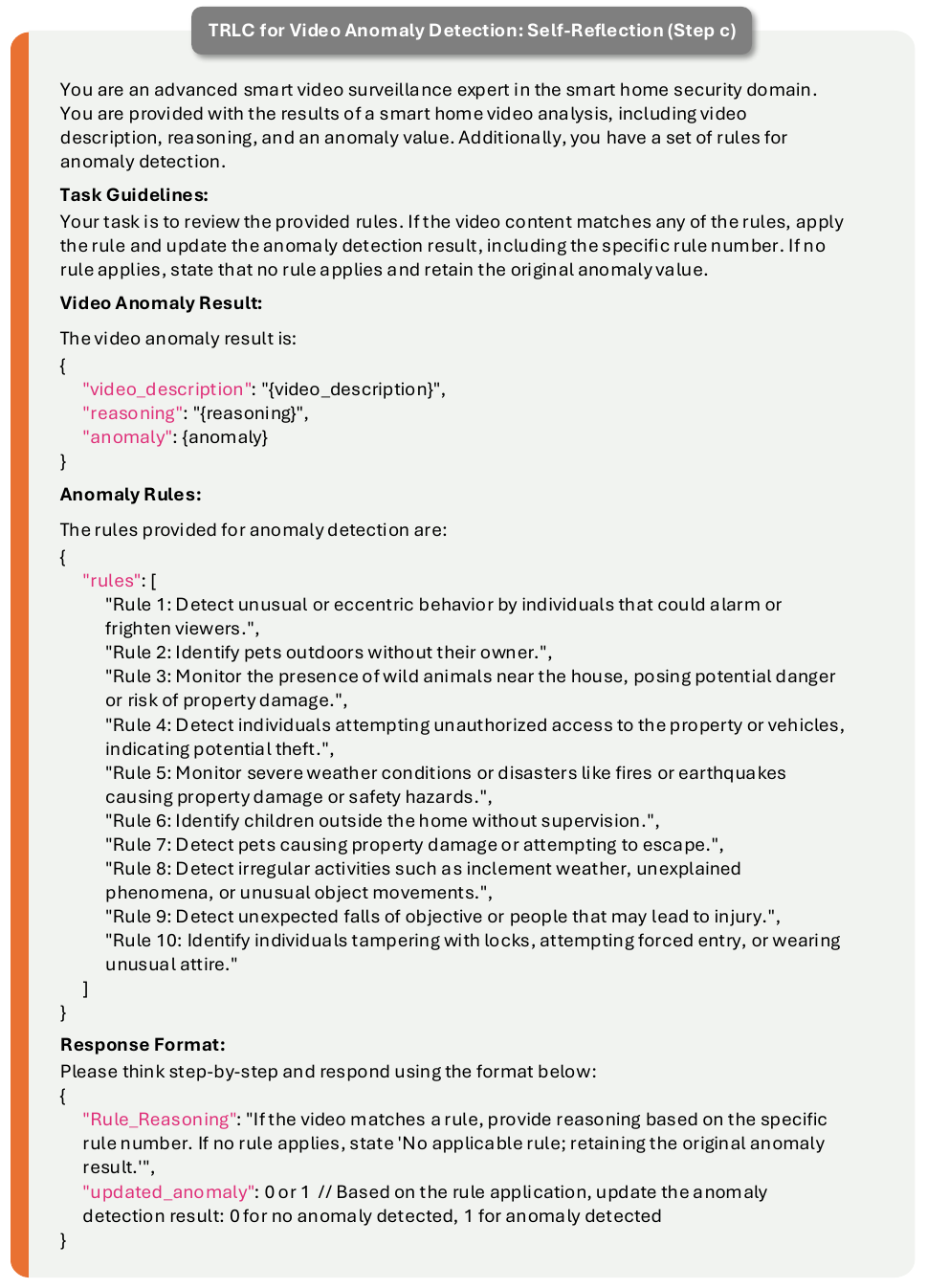}
\caption{System prompts adopted in step (c) of the TRLC for VAD: self-reflection.}\label{fig:chain-c-prompt}
\end{figure*}

The prompts used in the taxonomy-driven reflective LLM chain (TRLC) framework for the VAD task are detailed as follows. First, the prompts in Figure \ref{fig:chain-a-prompt} are used in step (a) of the TRLC to generate rules from the complete video anomaly taxonomy, with the resulting rules from step (a) shown in Figure \ref{fig:10rule}. Next, the prompts in Figure \ref{fig:chain-b-prompt} are employed to predict the initial detection for the VAD task. Finally, the self-reflection step is carried out using the prompts provided in Figure \ref{fig:chain-c-prompt}.


\newpage 

\newpage 

\subsection{System Prompt for Error Diagnosis in In-Depth Analysis}\label{appx:indepth_prompt}
\begin{figure*}[!ht]
\centering
\includegraphics[width=0.7\textwidth]{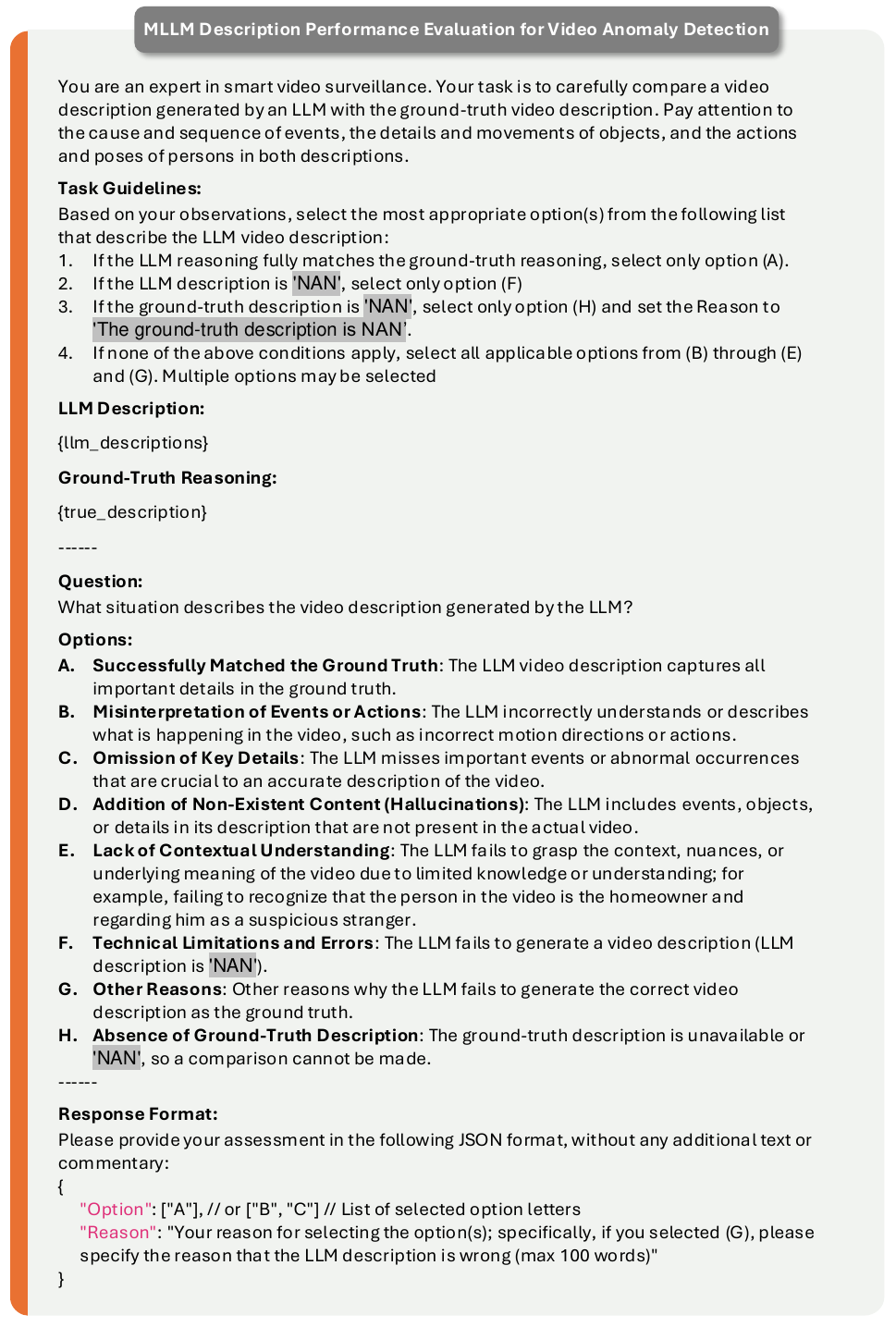}
\caption{System prompts adopted in evaluating the MLLM-generated video description for VAD.}\label{fig:des-prompt}
\end{figure*}

\begin{figure*}[!ht]
\centering
\includegraphics[width=0.7\textwidth]{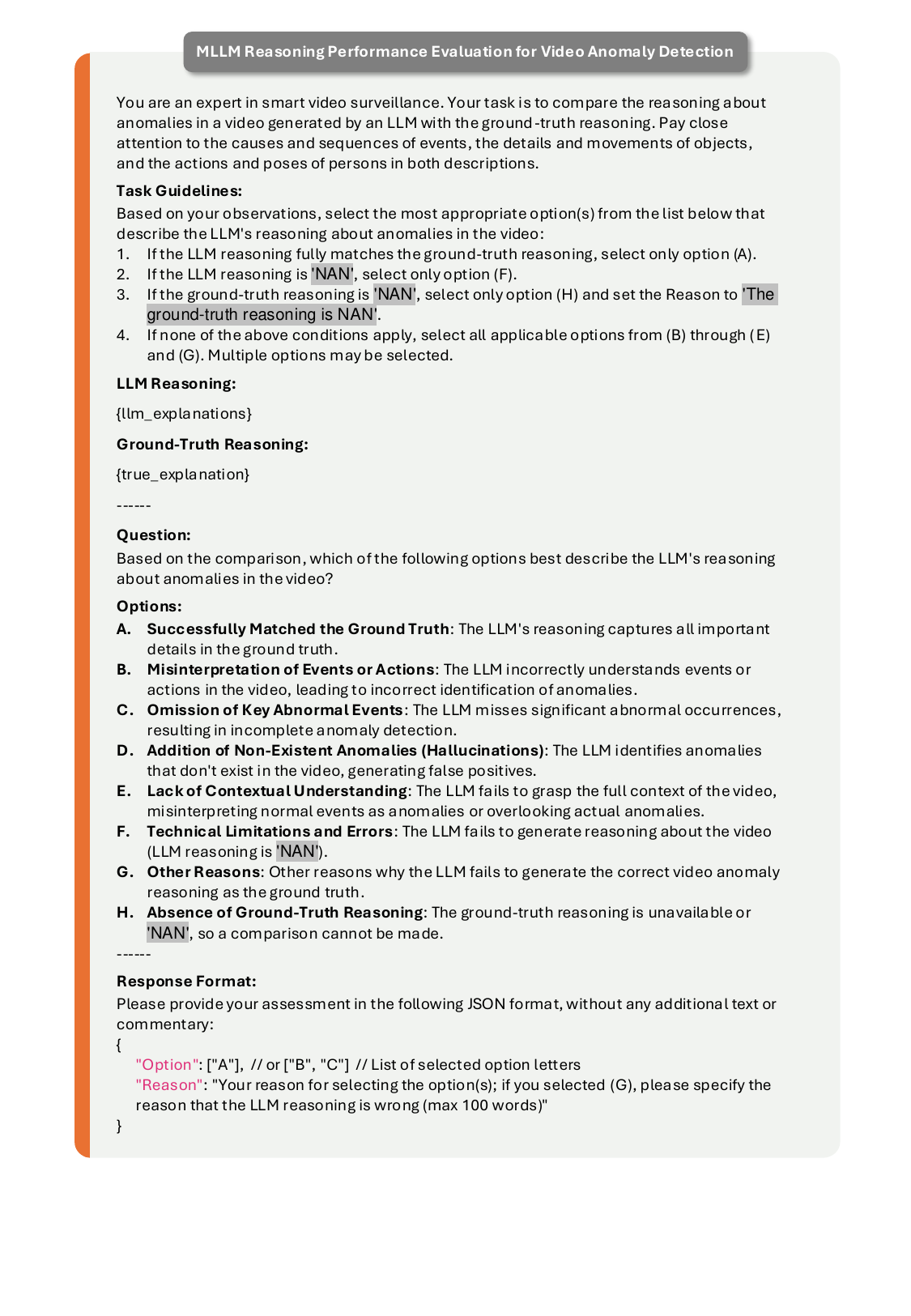}
\caption{System prompts adopted in evaluating the MLLM-generated video reasoning for VAD.}\label{fig:reason-prompt}
\end{figure*}

We use the prompts in Figure \ref{fig:des-prompt} and Figure \ref{fig:reason-prompt} to evaluate MLLM-generated video descriptions and reasoning against human-annotated counterparts, respectively.

\newpage 

\newpage

\section{Additional Experiments}\label{appx:results}
\subsection{Comparison between Anomaly Detection and Normality Detection}
Anomaly detection is a classical binary classification task \cite{bendou2024llm}. In the context of VAD, we employ two distinct prompt frames to evaluate the accuracy of MLLMs in this classification task. First, we prompt the MLLMs to identify abnormal events within a sequence of normal activities, targeting the anomaly detection task. Conversely, given that ``normal videos" constitute the majority of training data \cite{sun2019abnormal}, we also frame the task as a normality detection issue, prompting MLLMs to justify whether a video is normal. This bidirectional approach allows for a comprehensive evaluation of the MLLMs' capabilities in understanding and reasoning about smart home video clips, highlighting performance differences across different task frames in MLLM-based VAD.

\begin{figure*}[!ht]
\centering
\includegraphics[width=0.7\textwidth]{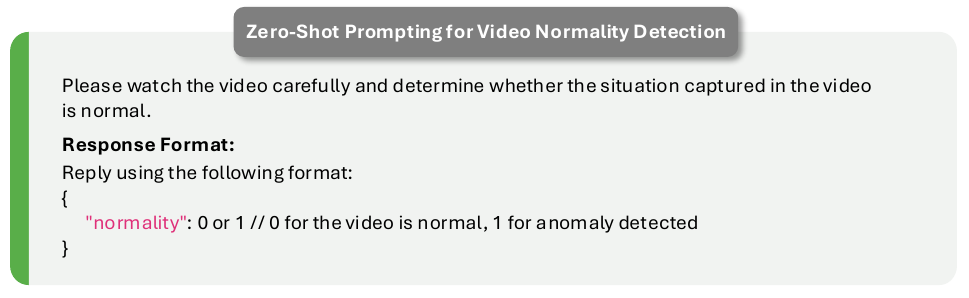}
\caption{System prompts adopted in zero-shot prompting for video normality detection.}\label{fig:0shotprompt-ND}
\end{figure*}

\paragraph{Zero-Shot Prompting} The zero-shot prompt for anomaly detection is illustrated in Figure \ref{fig:0shotprompt-AD}, while the prompt for normality detection is provided in Figure \ref{fig:0shotprompt-ND}. Table \ref{tab:0shot} presents the VAD results for both anomaly detection and normality detection tasks using zero-shot prompting. All MLLMs, except Claude-3.5-sonnet, achieve higher accuracy, precision, and recall in the normality detection task. VILA-13b classifies all videos as normal when tasked with anomaly detection, emphasizing its limitations in zero-shot VAD tasks, despite being the fastest model in processing videos. Given that VAD is a binary classification task, the random guess accuracy is 50\%. Even the best-performing MLLMs achieve accuracy close to this threshold, highlighting their limited understanding of anomalies in smart home contexts. These results likely reflect the models' training on datasets primarily composed of normal videos, leading to stronger prior knowledge of normal events in smart home scenarios.

\begin{table*}[!t]
  \centering
  \caption{Performance of MLLMs with two prompt frames: accuracy, precision, recall (\%), and processing time (s) compared across different MLLMs using zero-shot prompting (AD: anomaly detection, ND: normality detection).}
    \begin{tabular}{l*{8}{>{\centering\arraybackslash}m{1.4cm}}} 
    \toprule
    \multicolumn{1}{c}{\multirow{2}[4]{*}{\bf Model}} & \multicolumn{2}{c}{\bf Accuracy} & \multicolumn{2}{c}{\bf Precision} & \multicolumn{2}{c}{\bf Recall} & \multicolumn{2}{c}{\bf Video Processing Time} \\
\cmidrule{2-9}          & \bf AD & \bf ND & \bf AD & \bf ND & \bf AD & \bf ND & \bf AD & \bf ND \\
    \midrule
    Gemini-1.5-flash & 58.44 & 72.90 & 79.22 & 81.36 & 31.12 & 64.56 & 3.43  & 3.26 \\
    Gemini-1.5-pro & 57.36 & \textbf{74.15} & \textbf{84.34} & \textbf{86.58} & 25.73 & 61.63 & 4.14  & 4.02 \\
    GPT-4o & 68.41 & 70.74 & 80.09 & 82.07 & 55.16 & 58.55 & 10.15 & 9.79 \\
    GPT-4o-mini & 69.91 & 73.07 & 76.52 & 78.66 & 63.79 & \textbf{68.72} & 10.09 & 10.39 \\
    Claude-3.5-sonnet & \textbf{70.82} & 74.06 & 69.66 & 82.97 & \textbf{81.36} & 65.33 & 20.87 & 21.51 \\
    VILA-13b & 46.05 & 55.28 & 0.00  & 78.46 & 0.00  & 23.57 & \textbf{1.38}  & \textbf{1.28} \\
    \bottomrule
    \end{tabular}%
  \label{tab:0shot}%
\end{table*}%

\paragraph{CoT Prompting}
Given that all MLLMs perform better on normality detection than anomaly detection with zero-shot prompting, an important question arises: does this trend continue with CoT prompting?

\begin{figure*}[!ht]
\centering
\includegraphics[width=0.7\textwidth]{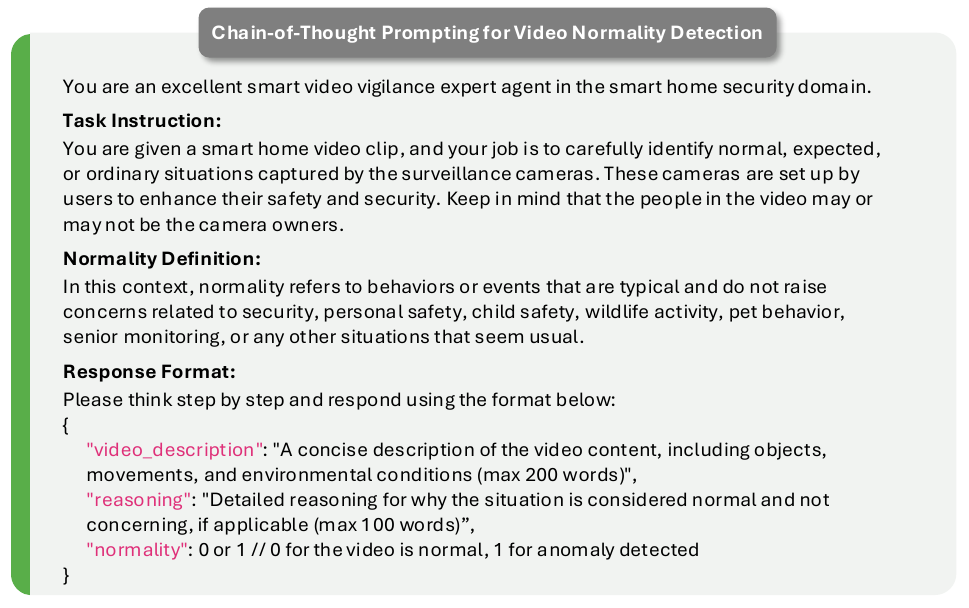}
\caption{System prompts adopted in CoT prompting for video normality detection.}\label{fig:cotprompt-ND}
\end{figure*} 

To investigate, we evaluate CoT performance for both anomaly detection and normality detection. The prompts used are detailed in Figure \ref{fig:cotprompt-AD} and Figure \ref{fig:cotprompt-ND}, respectively. As shown in Table \ref{tab:cot}, for the AD results, CoT prompting improves accuracy and recall compared to the zero-shot prompting in Table \ref{tab:0shot}, meeting expectations for CoT's effectiveness. However, performance in normality detection declines with CoT prompting. While four MLLMs achieve over 90\% precision in the normality detection task, the overall accuracy drops significantly compared to the ND results in Table \ref{tab:0shot}. This suggests that while MLLMs have a solid grasp of normality, CoT prompting reinforces their existing strengths without addressing their weaknesses in anomaly detection, resulting in a decrease in overall VAD accuracy. In terms of efficiency, Gemini-1.5-flash emerges as the fastest model with CoT prompting, whereas VILA-13b, previously the fastest, likely loses this advantage due to difficulties in processing longer prompts. 

\begin{table*}[!t]
  \centering
  \caption{
  Performance of MLLMs with two prompt frames: accuracy, precision, recall (\%), and processing time (s) compared across different MLLMs using CoT prompting (AD: anomaly detection, ND: normality detection).}
    \begin{tabular}{l*{8}{>{\centering\arraybackslash}m{1.4cm}}} 
    \toprule
    \multicolumn{1}{c}{\multirow{2}[4]{*}{\bf Model}} & \multicolumn{2}{c}{\bf Accuracy} & \multicolumn{2}{c}{\bf Precision} & \multicolumn{2}{c}{\bf Recall} & \multicolumn{2}{c}{\bf Video Processing Time} \\
\cmidrule{2-9}          & \bf AD & \bf ND & \bf AD & \bf ND & \bf AD & \bf ND & \bf AD & \bf ND \\
    \midrule
    Gemini-1.5-flash & 69.58 & 45.47 & 74.44 & 40.00 & 66.41 & 2.16  & \textbf{4.61}  & \textbf{4.57} \\
    Gemini-1.5-pro & \textbf{74.06} & \textbf{61.60} & \textbf{83.77} & 93.90 & 64.41 & \textbf{30.82} & 7.05  & 6.83 \\
    GPT-4o & 72.57 & 57.94 & 83.02 & \textbf{100.00} & 61.79 & 22.03 & 12.55 & 14.27 \\
    GPT-4o-mini & 68.83 & 49.46 & 68.07 & \textbf{100.00} & \textbf{79.51} & 6.32  & 12.28 & 13.39 \\
    Claude-3.5-sonnet & 71.90 & 54.20 & 83.44 & 95.37 & 59.78 & 15.87 & 24.49 & 24.09 \\
    VILA-13b & 68.41 & 43.39 & 68.45 & 13.64 & 76.89 & 0.92  & 6.74  & 11.56 \\
    \bottomrule
    \end{tabular}%
  \label{tab:cot}%
\end{table*}%


From the comparison between two prompt frames under zero-shot and CoT prompting, we observe that a feasible way to stably enhance MLLM VAD performance is to focus on anomaly detection while enriching the prompt with contextual information about anomalies in smart home scenarios. This strategy helps compensate for the models' inherent limited understanding of anomalies.

\subsection{Evaluation on Video Understanding of MLLMs}
\begin{figure*}[!h]
\centering
\includegraphics[width=\textwidth]{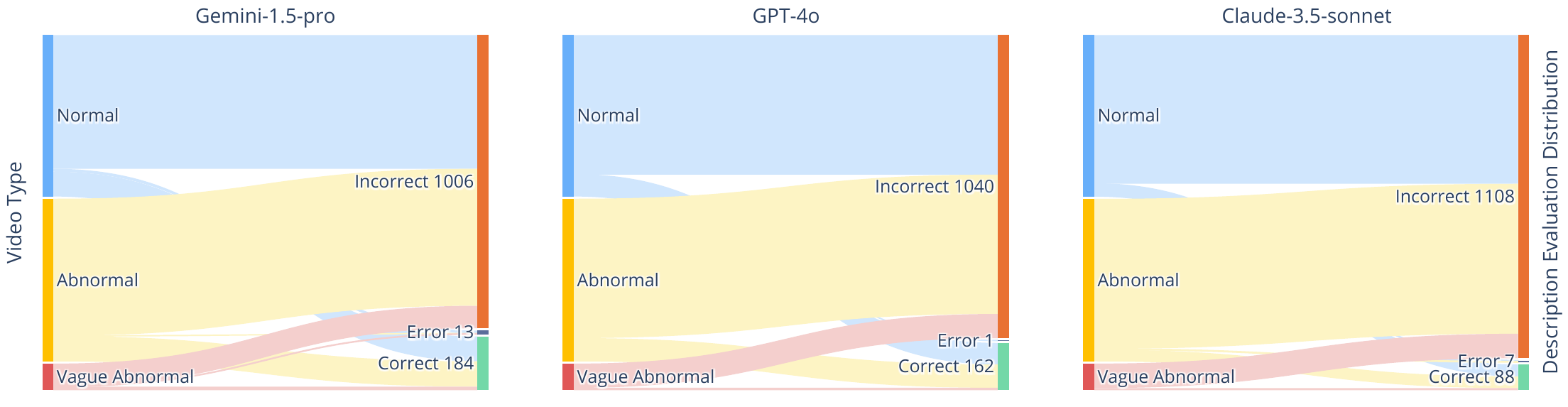}
\caption{Distribution of video outcomes for the top three MLLMs’ description compared to human-annotated description across different video anomaly tags.}\label{fig:sankeydes}
\end{figure*}

\begin{figure*}[!h]
\centering
\includegraphics[width=\textwidth]{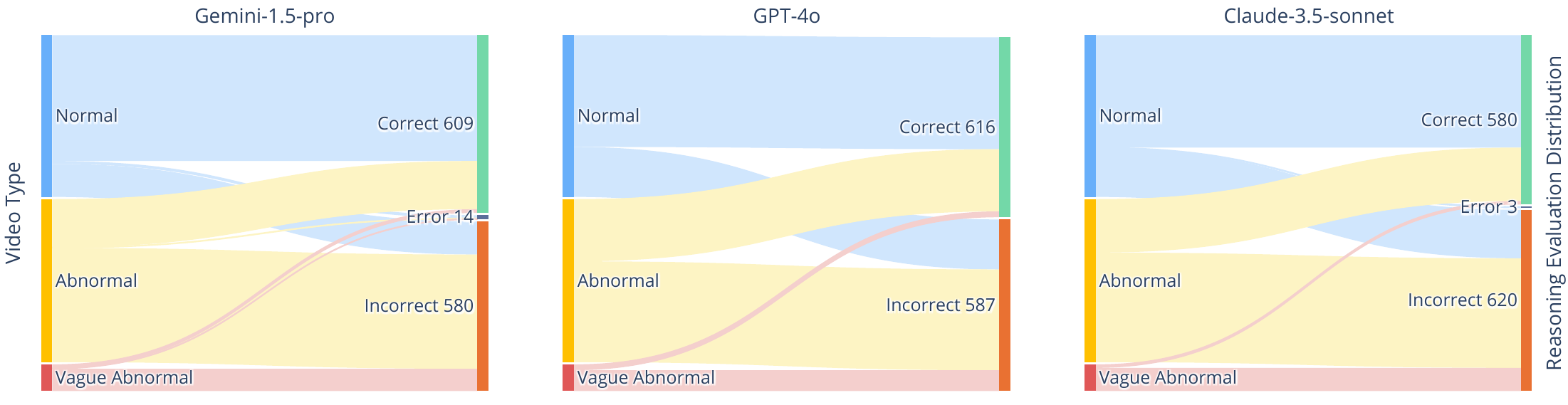}
\caption{Distribution of video outcomes for the top three MLLMs’ reasoning compared to human-annotated reasoning across different video anomaly tags.}\label{fig:sankeyres}
\end{figure*}

From Figure 7 and Figure 8 in the main paper, we analyze the five failure types where MLLMs failed to generate correct video description and reasoning. Additionally, we examine the distribution of MLLM outcomes for video description and reasoning across three ground-truth anomaly tags, i.e., \texttt{Normal}, \texttt{Abnormal}, and \texttt{Vague Abnormal}, as shown in Figures \ref{fig:sankeydes} and \ref{fig:sankeyres}, respectively. The possible outcomes are defined as follows: (1) Correct: the MLLM's response matches the annotated description or reasoning; (2) Error: the MLLM generates ``nan" or nonsensical information; (3) Incorrect: there is at least one mismatch between the MLLM output and human annotation.

For video description, over 1000 MLLM outputs are incorrect from the top three MLLMs, whereas over half of the reasoning outputs are correct. This discrepancy is likely because the description tends to include more detailed information compared to the reasoning, as illustrated in Figure \ref{fig:video_time_dis}, making it more challenging for MLLMs to match every detail in the descriptions. The error rates for the three models follow the same ranking for both description and reasoning: Gemini-1.5-pro exhibits the highest error rate, followed by Claude-3.5-sonnet, with GPT-4o showing the least, indicating the relative stability of GPT-4o in response generation. The proportion of videos with correct descriptions across MLLMs remains consistent between \texttt{normal} and \texttt{abnormal} videos. However, the proportion of correct reasoning decreases progressively from \texttt{normal} to \texttt{abnormal} and further to \texttt{vague abnormal}. This trend highlights the limited understanding MLLMs have of smart home anomalies in our dataset, particularly for more ambiguous cases.
\vspace{-1em}
\vspace{-1em}





%% file: main.bbl
\begin{thebibliography}{58}
\providecommand{\natexlab}[1]{#1}
\providecommand{\url}[1]{\texttt{#1}}
\expandafter\ifx\csname urlstyle\endcsname\relax
  \providecommand{\doi}[1]{doi: #1}\else
  \providecommand{\doi}{doi: \begingroup \urlstyle{rm}\Url}\fi

\bibitem[Achiam et~al.(2023)Achiam, Adler, Agarwal, Ahmad, Akkaya, Aleman, Almeida, Altenschmidt, Altman, Anadkat, et~al.]{achiam2023gpt}
Josh Achiam, Steven Adler, Sandhini Agarwal, Lama Ahmad, Ilge Akkaya, Florencia~Leoni Aleman, Diogo Almeida, Janko Altenschmidt, Sam Altman, Shyamal Anadkat, et~al.
\newblock Gpt-4 technical report.
\newblock \emph{arXiv preprint arXiv:2303.08774}, 2023.

\bibitem[Alayrac et~al.(2022)Alayrac, Donahue, Luc, Miech, Barr, Hasson, Lenc, Mensch, Millican, Reynolds, et~al.]{alayrac2022flamingo}
Jean-Baptiste Alayrac, Jeff Donahue, Pauline Luc, Antoine Miech, Iain Barr, Yana Hasson, Karel Lenc, Arthur Mensch, Katherine Millican, Malcolm Reynolds, et~al.
\newblock Flamingo: a visual language model for few-shot learning.
\newblock \emph{Advances in neural information processing systems}, 35:\penalty0 23716--23736, 2022.

\bibitem[Ali(2023)]{ali2023real}
Manal~Mostafa Ali.
\newblock Real-time video anomaly detection for smart surveillance.
\newblock \emph{IET Image Processing}, 17\penalty0 (5):\penalty0 1375--1388, 2023.

\bibitem[Anthropic(2024)]{anthropic2024claude3sonnet}
Anthropic.
\newblock Claude 3.5 sonnet, 2024.
\newblock Accessed: 2025-04-05.

\bibitem[Bakar et~al.(2015)Bakar, Ghayvat, Hasanm, and Mukhopadhyay]{bakar2015activity}
UABUA Bakar, Hemant Ghayvat, SF Hasanm, and Subhas~Chandra Mukhopadhyay.
\newblock Activity and anomaly detection in smart home: A survey.
\newblock \emph{Next generation sensors and systems}, pages 191--220, 2015.

\bibitem[Bendou et~al.(2024)Bendou, Lioi, Pasdeloup, Mauch, Hacene, Cardinaux, and Gripon]{bendou2024llm}
Yassir Bendou, Giulia Lioi, Bastien Pasdeloup, Lukas Mauch, Ghouthi~Boukli Hacene, Fabien Cardinaux, and Vincent Gripon.
\newblock Llm meets vision-language models for zero-shot one-class classification.
\newblock \emph{arXiv preprint arXiv:2404.00675}, 2024.

\bibitem[Bharadwaj et~al.(2024)Bharadwaj, Gani, Naseer, Khan, and Khan]{bharadwaj2024vane}
Rohit Bharadwaj, Hanan Gani, Muzammal Naseer, Fahad~Shahbaz Khan, and Salman Khan.
\newblock Vane-bench: Video anomaly evaluation benchmark for conversational lmms.
\newblock \emph{arXiv preprint arXiv:2406.10326}, 2024.

\bibitem[Corona et~al.(2021)Corona, Osterdahl, Collins, and Hoogs]{corona2021meva}
Kellie Corona, Katie Osterdahl, Roderic Collins, and Anthony Hoogs.
\newblock Meva: A large-scale multiview, multimodal video dataset for activity detection.
\newblock In \emph{Proceedings of the IEEE/CVF winter conference on applications of computer vision}, pages 1060--1068, 2021.

\bibitem[Driess et~al.(2023)Driess, Xia, Sajjadi, Lynch, Chowdhery, Ichter, Wahid, Tompson, Vuong, Yu, et~al.]{driess2023palm}
Danny Driess, Fei Xia, Mehdi~SM Sajjadi, Corey Lynch, Aakanksha Chowdhery, Brian Ichter, Ayzaan Wahid, Jonathan Tompson, Quan Vuong, Tianhe Yu, et~al.
\newblock Palm-e: An embodied multimodal language model.
\newblock \emph{arXiv preprint arXiv:2303.03378}, 2023.

\bibitem[Fu et~al.(2024)Fu, Dai, Luo, Li, Ren, Zhang, Wang, Zhou, Shen, Zhang, et~al.]{fu2024video}
Chaoyou Fu, Yuhan Dai, Yondong Luo, Lei Li, Shuhuai Ren, Renrui Zhang, Zihan Wang, Chenyu Zhou, Yunhang Shen, Mengdan Zhang, et~al.
\newblock Video-mme: The first-ever comprehensive evaluation benchmark of multi-modal llms in video analysis.
\newblock \emph{arXiv preprint arXiv:2405.21075}, 2024.

\bibitem[Gong et~al.(2019)Gong, Liu, Le, Saha, Mansour, Venkatesh, and Hengel]{gong2019memorizing}
Dong Gong, Lingqiao Liu, Vuong Le, Budhaditya Saha, Moussa~Reda Mansour, Svetha Venkatesh, and Anton van~den Hengel.
\newblock Memorizing normality to detect anomaly: Memory-augmented deep autoencoder for unsupervised anomaly detection.
\newblock In \emph{Proceedings of the IEEE/CVF international conference on computer vision}, pages 1705--1714, 2019.

\bibitem[Hasan et~al.(2016)Hasan, Choi, Neumann, Roy-Chowdhury, and Davis]{hasan2016learning}
Mahmudul Hasan, Jonghyun Choi, Jan Neumann, Amit~K Roy-Chowdhury, and Larry~S Davis.
\newblock Learning temporal regularity in video sequences.
\newblock In \emph{Proceedings of the IEEE conference on computer vision and pattern recognition}, pages 733--742, 2016.

\bibitem[Hurst et~al.(2024)Hurst, Lerer, Goucher, Perelman, Ramesh, Clark, Ostrow, Welihinda, Hayes, Radford, et~al.]{hurst2024gpt}
Aaron Hurst, Adam Lerer, Adam~P Goucher, Adam Perelman, Aditya Ramesh, Aidan Clark, AJ Ostrow, Akila Welihinda, Alan Hayes, Alec Radford, et~al.
\newblock Gpt-4o system card.
\newblock \emph{arXiv preprint arXiv:2410.21276}, 2024.

\bibitem[Inan et~al.(2023)Inan, Upasani, Chi, Rungta, Iyer, Mao, Tontchev, Hu, Fuller, Testuggine, et~al.]{inan2023llama}
Hakan Inan, Kartikeya Upasani, Jianfeng Chi, Rashi Rungta, Krithika Iyer, Yuning Mao, Michael Tontchev, Qing Hu, Brian Fuller, Davide Testuggine, et~al.
\newblock Llama guard: Llm-based input-output safeguard for human-ai conversations.
\newblock \emph{arXiv preprint arXiv:2312.06674}, 2023.

\bibitem[Ionescu et~al.(2019)Ionescu, Khan, Georgescu, and Shao]{ionescu2019object}
Radu~Tudor Ionescu, Fahad~Shahbaz Khan, Mariana-Iuliana Georgescu, and Ling Shao.
\newblock Object-centric auto-encoders and dummy anomalies for abnormal event detection in video.
\newblock In \emph{Proceedings of the IEEE/CVF conference on computer vision and pattern recognition}, pages 7842--7851, 2019.

\bibitem[Kim and Moon(2022)]{kim2022dog}
Jinah Kim and Nammee Moon.
\newblock Dog behavior recognition based on multimodal data from a camera and wearable device.
\newblock \emph{Applied sciences}, 12\penalty0 (6):\penalty0 3199, 2022.

\bibitem[Li et~al.(2023{\natexlab{a}})Li, Li, Savarese, and Hoi]{li2023blip}
Junnan Li, Dongxu Li, Silvio Savarese, and Steven Hoi.
\newblock Blip-2: Bootstrapping language-image pre-training with frozen image encoders and large language models.
\newblock In \emph{International conference on machine learning}, pages 19730--19742. PMLR, 2023{\natexlab{a}}.

\bibitem[Li et~al.(2023{\natexlab{b}})Li, He, Wang, Li, Wang, Luo, Wang, Wang, and Qiao]{li2023videochat}
KunChang Li, Yinan He, Yi Wang, Yizhuo Li, Wenhai Wang, Ping Luo, Yali Wang, Limin Wang, and Yu Qiao.
\newblock Videochat: Chat-centric video understanding.
\newblock \emph{arXiv preprint arXiv:2305.06355}, 2023{\natexlab{b}}.

\bibitem[Li et~al.(2024{\natexlab{a}})Li, Wang, He, Li, Wang, Liu, Wang, Xu, Chen, Luo, et~al.]{li2024mvbench}
Kunchang Li, Yali Wang, Yinan He, Yizhuo Li, Yi Wang, Yi Liu, Zun Wang, Jilan Xu, Guo Chen, Ping Luo, et~al.
\newblock Mvbench: A comprehensive multi-modal video understanding benchmark.
\newblock In \emph{Proceedings of the IEEE/CVF Conference on Computer Vision and Pattern Recognition}, pages 22195--22206, 2024{\natexlab{a}}.

\bibitem[Li et~al.(2022)Li, Liu, and Jiao]{li2022self}
Shuo Li, Fang Liu, and Licheng Jiao.
\newblock Self-training multi-sequence learning with transformer for weakly supervised video anomaly detection.
\newblock In \emph{Proceedings of the AAAI Conference on Artificial Intelligence}, pages 1395--1403, 2022.

\bibitem[Li et~al.(2013)Li, Mahadevan, and Vasconcelos]{li2013anomaly}
Weixin Li, Vijay Mahadevan, and Nuno Vasconcelos.
\newblock Anomaly detection and localization in crowded scenes.
\newblock \emph{IEEE transactions on pattern analysis and machine intelligence}, 36\penalty0 (1):\penalty0 18--32, 2013.

\bibitem[Li et~al.(2024{\natexlab{b}})Li, Chen, Hu, Wang, Shi, and Zhang]{li2024videovista}
Yunxin Li, Xinyu Chen, Baotian Hu, Longyue Wang, Haoyuan Shi, and Min Zhang.
\newblock Videovista: A versatile benchmark for video understanding and reasoning.
\newblock \emph{arXiv preprint arXiv:2406.11303}, 2024{\natexlab{b}}.

\bibitem[Lin et~al.(2024)Lin, Yin, Ping, Molchanov, Shoeybi, and Han]{lin2024vila}
Ji Lin, Hongxu Yin, Wei Ping, Pavlo Molchanov, Mohammad Shoeybi, and Song Han.
\newblock Vila: On pre-training for visual language models.
\newblock In \emph{Proceedings of the IEEE/CVF Conference on Computer Vision and Pattern Recognition}, pages 26689--26699, 2024.

\bibitem[Liu et~al.(2021)Liu, Xia, and Tang]{liu2021privacy}
Jixin Liu, Yinyun Xia, and Zheng Tang.
\newblock Privacy-preserving video fall detection using visual shielding information.
\newblock \emph{The Visual Computer}, 37\penalty0 (2):\penalty0 359--370, 2021.

\bibitem[Liu et~al.(2018)Liu, Luo, Lian, and Gao]{liu2018future}
Wen Liu, Weixin Luo, Dongze Lian, and Shenghua Gao.
\newblock Future frame prediction for anomaly detection--a new baseline.
\newblock In \emph{Proceedings of the IEEE conference on computer vision and pattern recognition}, pages 6536--6545, 2018.

\bibitem[Liu et~al.(2024)Liu, Li, Liu, Wang, Ren, Li, Chen, Sun, and Hou]{liu2024tempcompass}
Yuanxin Liu, Shicheng Li, Yi Liu, Yuxiang Wang, Shuhuai Ren, Lei Li, Sishuo Chen, Xu Sun, and Lu Hou.
\newblock Tempcompass: Do video llms really understand videos?
\newblock \emph{arXiv preprint arXiv:2403.00476}, 2024.

\bibitem[Lu et~al.(2013)Lu, Shi, and Jia]{lu2013abnormal}
Cewu Lu, Jianping Shi, and Jiaya Jia.
\newblock Abnormal event detection at 150 fps in matlab.
\newblock In \emph{Proceedings of the IEEE international conference on computer vision}, pages 2720--2727, 2013.

\bibitem[Lv and Sun(2024)]{lv2024video}
Hui Lv and Qianru Sun.
\newblock Video anomaly detection and explanation via large language models.
\newblock \emph{arXiv preprint arXiv:2401.05702}, 2024.

\bibitem[Lv et~al.(2021)Lv, Chen, Cui, Xu, Li, and Yang]{lv2021learning}
Hui Lv, Chen Chen, Zhen Cui, Chunyan Xu, Yong Li, and Jian Yang.
\newblock Learning normal dynamics in videos with meta prototype network.
\newblock In \emph{Proceedings of the IEEE/CVF conference on computer vision and pattern recognition}, pages 15425--15434, 2021.

\bibitem[Maaz et~al.(2023)Maaz, Rasheed, Khan, and Khan]{maaz2023video}
Muhammad Maaz, Hanoona Rasheed, Salman Khan, and Fahad~Shahbaz Khan.
\newblock Video-chatgpt: Towards detailed video understanding via large vision and language models.
\newblock \emph{arXiv preprint arXiv:2306.05424}, 2023.

\bibitem[Markovitz et~al.(2020)Markovitz, Sharir, Friedman, Zelnik-Manor, and Avidan]{markovitz2020graph}
Amir Markovitz, Gilad Sharir, Itamar Friedman, Lihi Zelnik-Manor, and Shai Avidan.
\newblock Graph embedded pose clustering for anomaly detection.
\newblock In \emph{Proceedings of the IEEE/CVF Conference on Computer Vision and Pattern Recognition}, pages 10539--10547, 2020.

\bibitem[Nam et~al.(2024)Nam, Macvean, Hellendoorn, Vasilescu, and Myers]{nam2024using}
Daye Nam, Andrew Macvean, Vincent Hellendoorn, Bogdan Vasilescu, and Brad Myers.
\newblock Using an llm to help with code understanding.
\newblock In \emph{Proceedings of the IEEE/ACM 46th International Conference on Software Engineering}, pages 1--13, 2024.

\bibitem[Nayak et~al.(2021)Nayak, Pati, and Das]{nayak2021comprehensive}
Rashmiranjan Nayak, Umesh~Chandra Pati, and Santos~Kumar Das.
\newblock A comprehensive review on deep learning-based methods for video anomaly detection.
\newblock \emph{Image and Vision Computing}, 106:\penalty0 104078, 2021.

\bibitem[Ntelopoulos and Nasrollahi(2024)]{ntelopoulos2024callm}
Apostolos Ntelopoulos and Kamal Nasrollahi.
\newblock Callm: Cascading autoencoder and large language model for video anomaly detection.
\newblock In \emph{International Conference on Image Processing Theory, Tools and Applications}. IEEE, 2024.

\bibitem[Oelschlager(2024)]{oelschlager2024evaluating}
Richard Oelschlager.
\newblock Evaluating the impact of hallucinations on user trust and satisfaction in llm-based systems, 2024.

\bibitem[OpenAI(2024)]{openai2023gpt4omini}
OpenAI.
\newblock Gpt-4o-mini: Advancing cost-efficient intelligence, 2024.
\newblock Accessed: 2025-04-05.

\bibitem[Pandya et~al.(2018)Pandya, Ghayvat, Kotecha, Awais, Akbarzadeh, Gope, Mukhopadhyay, and Chen]{pandya2018smart}
Sharnil Pandya, Hemant Ghayvat, Ketan Kotecha, Mohammed Awais, Saeed Akbarzadeh, Prosanta Gope, Subhas~Chandra Mukhopadhyay, and Wei Chen.
\newblock Smart home anti-theft system: a novel approach for near real-time monitoring and smart home security for wellness protocol.
\newblock \emph{Applied System Innovation}, 1\penalty0 (4):\penalty0 42, 2018.

\bibitem[Ren et~al.(2021)Ren, Xia, Liu, and Lee]{ren2021deep}
Jing Ren, Feng Xia, Yemeng Liu, and Ivan Lee.
\newblock Deep video anomaly detection: Opportunities and challenges.
\newblock In \emph{2021 international conference on data mining workshops (ICDMW)}, pages 959--966. IEEE, 2021.

\bibitem[Sultani et~al.(2018)Sultani, Chen, and Shah]{sultani2018real}
Waqas Sultani, Chen Chen, and Mubarak Shah.
\newblock Real-world anomaly detection in surveillance videos.
\newblock In \emph{Proceedings of the IEEE conference on computer vision and pattern recognition}, pages 6479--6488, 2018.

\bibitem[Sun et~al.(2019)Sun, Shao, and He]{sun2019abnormal}
Jiayu Sun, Jie Shao, and Chengkun He.
\newblock Abnormal event detection for video surveillance using deep one-class learning.
\newblock \emph{Multimedia Tools and Applications}, 78\penalty0 (3):\penalty0 3633--3647, 2019.

\bibitem[Team et~al.(2023)Team, Anil, Borgeaud, Wu, Alayrac, Yu, Soricut, Schalkwyk, Dai, Hauth, et~al.]{team2023gemini}
Gemini Team, Rohan Anil, Sebastian Borgeaud, Yonghui Wu, Jean-Baptiste Alayrac, Jiahui Yu, Radu Soricut, Johan Schalkwyk, Andrew~M Dai, Anja Hauth, et~al.
\newblock Gemini: a family of highly capable multimodal models.
\newblock \emph{arXiv preprint arXiv:2312.11805}, 2023.

\bibitem[Team et~al.(2024)Team, Georgiev, Lei, Burnell, Bai, Gulati, Tanzer, Vincent, Pan, Wang, et~al.]{team2024gemini}
Gemini Team, Petko Georgiev, Ving~Ian Lei, Ryan Burnell, Libin Bai, Anmol Gulati, Garrett Tanzer, Damien Vincent, Zhufeng Pan, Shibo Wang, et~al.
\newblock Gemini 1.5: Unlocking multimodal understanding across millions of tokens of context.
\newblock \emph{arXiv preprint arXiv:2403.05530}, 2024.

\bibitem[Tian et~al.(2021)Tian, Pang, Chen, Singh, Verjans, and Carneiro]{tian2021weakly}
Yu Tian, Guansong Pang, Yuanhong Chen, Rajvinder Singh, Johan~W Verjans, and Gustavo Carneiro.
\newblock Weakly-supervised video anomaly detection with robust temporal feature magnitude learning.
\newblock In \emph{Proceedings of the IEEE/CVF international conference on computer vision}, pages 4975--4986, 2021.

\bibitem[Wei et~al.(2022)Wei, Wang, Schuurmans, Bosma, Xia, Chi, Le, Zhou, et~al.]{wei2022chain}
Jason Wei, Xuezhi Wang, Dale Schuurmans, Maarten Bosma, Fei Xia, Ed Chi, Quoc~V Le, Denny Zhou, et~al.
\newblock Chain-of-thought prompting elicits reasoning in large language models.
\newblock \emph{Advances in neural information processing systems}, 35:\penalty0 24824--24837, 2022.

\bibitem[Withanage et~al.(2016)Withanage, Lee, Brinkworth, Mackintosh, and Thewlis]{withanage2016fall}
Kalana~Ishara Withanage, Ivan Lee, Russell Brinkworth, Shylie Mackintosh, and Dominic Thewlis.
\newblock Fall recovery subactivity recognition with rgb-d cameras.
\newblock \emph{IEEE transactions on industrial informatics}, 12\penalty0 (6):\penalty0 2312--2320, 2016.

\bibitem[Wu and Liu(2021)]{wu2021learning}
Peng Wu and Jing Liu.
\newblock Learning causal temporal relation and feature discrimination for anomaly detection.
\newblock \emph{IEEE Transactions on Image Processing}, 30:\penalty0 3513--3527, 2021.

\bibitem[Wu et~al.(2020)Wu, Liu, Shi, Sun, Shao, Wu, and Yang]{wu2020not}
Peng Wu, Jing Liu, Yujia Shi, Yujia Sun, Fangtao Shao, Zhaoyang Wu, and Zhiwei Yang.
\newblock Not only look, but also listen: Learning multimodal violence detection under weak supervision.
\newblock In \emph{Computer Vision--ECCV 2020: 16th European Conference, Glasgow, UK, August 23--28, 2020, Proceedings, Part XXX 16}, pages 322--339. Springer, 2020.

\bibitem[Wu et~al.(2022)Wu, Terry, and Cai]{wu2022ai}
Tongshuang Wu, Michael Terry, and Carrie~Jun Cai.
\newblock Ai chains: Transparent and controllable human-ai interaction by chaining large language model prompts.
\newblock In \emph{Proceedings of the 2022 CHI conference on human factors in computing systems}, pages 1--22, 2022.

\bibitem[Xiao et~al.(2021)Xiao, Shang, Yao, and Chua]{xiao2021next}
Junbin Xiao, Xindi Shang, Angela Yao, and Tat-Seng Chua.
\newblock Next-qa: Next phase of question-answering to explaining temporal actions.
\newblock In \emph{Proceedings of the IEEE/CVF conference on computer vision and pattern recognition}, pages 9777--9786, 2021.

\bibitem[Xu et~al.(2024)Xu, Cao, Chen, Shen, and Huang]{xu2024customizing}
Xiaohao Xu, Yunkang Cao, Yongqi Chen, Weiming Shen, and Xiaonan Huang.
\newblock Customizing visual-language foundation models for multi-modal anomaly detection and reasoning.
\newblock \emph{arXiv preprint arXiv:2403.11083}, 2024.

\bibitem[Yahaya et~al.(2021)Yahaya, Lotfi, and Mahmud]{yahaya2021towards}
Salisu~Wada Yahaya, Ahmad Lotfi, and Mufti Mahmud.
\newblock Towards a data-driven adaptive anomaly detection system for human activity.
\newblock \emph{Pattern Recognition Letters}, 145:\penalty0 200--207, 2021.

\bibitem[Yamauchi et~al.(2020)Yamauchi, Ohsita, Murata, Ueda, and Kato]{yamauchi2020anomaly}
Masaaki Yamauchi, Yuichi Ohsita, Masayuki Murata, Kensuke Ueda, and Yoshiaki Kato.
\newblock Anomaly detection in smart home operation from user behaviors and home conditions.
\newblock \emph{IEEE Transactions on Consumer Electronics}, 66\penalty0 (2):\penalty0 183--192, 2020.

\bibitem[Yang et~al.(2024)Yang, Lee, Dariush, Cao, and Lo]{yang2024follow}
Yuchen Yang, Kwonjoon Lee, Behzad Dariush, Yinzhi Cao, and Shao-Yuan Lo.
\newblock Follow the rules: Reasoning for video anomaly detection with large language models.
\newblock \emph{arXiv preprint arXiv:2407.10299}, 2024.

\bibitem[Zaheer et~al.(2022)Zaheer, Mahmood, Khan, Segu, Yu, and Lee]{zaheer2022generative}
M~Zaigham Zaheer, Arif Mahmood, M~Haris Khan, Mattia Segu, Fisher Yu, and Seung-Ik Lee.
\newblock Generative cooperative learning for unsupervised video anomaly detection.
\newblock In \emph{Proceedings of the IEEE/CVF conference on computer vision and pattern recognition}, pages 14744--14754, 2022.

\bibitem[Zanella et~al.(2024)Zanella, Menapace, Mancini, Wang, and Ricci]{zanella2024harnessing}
Luca Zanella, Willi Menapace, Massimiliano Mancini, Yiming Wang, and Elisa Ricci.
\newblock Harnessing large language models for training-free video anomaly detection.
\newblock In \emph{Proceedings of the IEEE/CVF Conference on Computer Vision and Pattern Recognition}, pages 18527--18536, 2024.

\bibitem[Zhang et~al.(2024)Zhang, Xu, Wang, Zuo, Han, Huang, Gao, Wang, and Sang]{zhang2024holmes}
Huaxin Zhang, Xiaohao Xu, Xiang Wang, Jialong Zuo, Chuchu Han, Xiaonan Huang, Changxin Gao, Yuehuan Wang, and Nong Sang.
\newblock Holmes-vad: Towards unbiased and explainable video anomaly detection via multi-modal llm.
\newblock \emph{arXiv preprint arXiv:2406.12235}, 2024.

\bibitem[Zhang et~al.(2015)Zhang, Shan, and Huang]{zhang2015isee}
Junge Zhang, Yanhu Shan, and Kaiqi Huang.
\newblock Isee smart home (ish): Smart video analysis for home security.
\newblock \emph{Neurocomputing}, 149:\penalty0 752--766, 2015.

\bibitem[Zhu et~al.(2021)Zhu, Chen, and Sultani]{zhu2021video}
Sijie Zhu, Chen Chen, and Waqas Sultani.
\newblock Video anomaly detection for smart surveillance.
\newblock In \emph{Computer Vision: A Reference Guide}, pages 1315--1322. Springer, 2021.

\end{thebibliography}
